\documentclass[12pt]{article}
\usepackage{amsmath}
\usepackage{times}
\usepackage{graphicx}
\usepackage{color}
\usepackage{multirow}
\usepackage[authoryear]{natbib}
\usepackage{rotating}
\usepackage{bbm}
\usepackage{latexsym}
\usepackage{subfigure}
\usepackage{booktabs} 
\usepackage{amssymb,amsthm}
\usepackage{microtype}
\usepackage{bibunits}
\usepackage{colortbl}
\usepackage{xcolor}
\usepackage{pifont}
\usepackage{footnote}

\usepackage{multirow,tabularx}
\usepackage{multicol}
\usepackage{comment}
\usepackage{mathtools}
\usepackage{algorithm,algcompatible}
\usepackage{caption}
\usepackage{wrapfig}
\usepackage{placeins}
\usepackage{blindtext}
\usepackage[english]{babel}

\newtheorem{corollary}{Corollary}
\newtheorem{theorem}{Theorem}
\newtheorem{lemma}{Lemma}

\usepackage{hyperref}

\newcommand{\eqann}[2][=]{\overset{\mathclap{(\text{#2})}}{#1}}
\newcommand{\eqannref}[1]{$(\text{#1})$}
\newcommand{\bv}[1]{\mathbf{#1}} 
\newcommand{\bs}[1]{\boldsymbol{#1}} 

\DeclareMathOperator{\Vectorization}{vec}
\DeclareMathOperator{\trace}{Tr}
\DeclareMathOperator{\kl}{\mathcal{D}_{\mathrm{KL}}}

\newcommand{\captionfonts}{\normalsize}

\makeatletter  
\long\def\@makecaption#1#2{%
  \vskip\abovecaptionskip
  \sbox\@tempboxa{{\captionfonts #1: #2}}%
  \ifdim \wd\@tempboxa >\hsize
    {\captionfonts #1: #2\par}
  \else
    \hbox to\hsize{\hfil\box\@tempboxa\hfil}%
  \fi
  \vskip\belowcaptionskip}
\makeatother   

\begin{document}
\hspace{13.9cm}1

\ \vspace{20mm}\\

{\LARGE Task Agnostic Continual Learning Using Online Variational Bayes with Fixed-Point Updates}

\ \\
{\bf \large Chen Zeno$^{\displaystyle 1}$, Itay Golan$^{\displaystyle 1}$, Elad Hoffer $^{\displaystyle 2}$ , Daniel Soudry $^{\displaystyle 1}$}\\
{$^{\displaystyle 1}$Department of Electrical Engineering, Technion, Israel Institute of Technology, Haifa, Israel.}\\
{$^{\displaystyle 2}$Habana-Labs, Caesarea, Israel.}\\
%

{\bf Keywords:} continual learning, variational Bayes, catastrophic forgetting, neural networks.

\thispagestyle{empty}
\markboth{}{NC instructions}
\ \vspace{-0mm}\\
%
\begin{center} {\bf Abstract} \end{center}
\textbf{Background:} Catastrophic forgetting is the notorious vulnerability of neural networks to the changes in the data distribution during learning. This phenomenon has long been considered a major obstacle for using learning agents in realistic continual learning settings. A large body of continual learning research assumes that task boundaries are known during training. However, only a few works consider scenarios in which task boundaries are unknown or not well defined --- task agnostic scenarios. The optimal Bayesian solution for this requires an intractable online Bayes update to the weights posterior. \newline
\textbf{Contributions:} We aim to approximate the online Bayes update as accurately as possible. To do so, we derive novel fixed-point equations for the online variational Bayes optimization problem, for multivariate Gaussian parametric distributions. By iterating the posterior through these fixed-point equations, we obtain an algorithm (FOO-VB) for continual learning which can handle non-stationary data distribution using a fixed architecture and without using external memory (i.e. without access to previous data). We demonstrate that our method (FOO-VB) outperforms existing methods in task agnostic scenarios. FOO-VB Pytorch implementation is available at \href{https://github.com/chenzeno/FOO-VB}{https://github.com/chenzeno/FOO-VB}.
\section{Introduction}

Continual learning (CL) is the ability of an algorithm to learn from non-stationary data while reusing past knowledge and exploiting it to better adapt to a changing environment. A major challenge in continual learning is to overcome 
\textit{catastrophic forgetting}~\citep{mccloskey1989catastrophic}. Catastrophic forgetting is the tendency of neural networks to rapidly lose previously learned knowledge when the input distribution is changed abruptly, e.g. when changing a task or a source.

Neural networks are commonly used machine learning models for solving a variety of tasks. They are notorious for being vulnerable to changes in the data during learning. Various methods for preventing catastrophic forgetting in neural networks have been suggested in the literature. Most of these methods assume relaxed conditions, in which the tasks arrive sequentially, and the data distribution changes only upon task-switches (i.e. piecewise stationary). Therefore, they are inapplicable in many realistic \textit{task agnostic} applications, in which the tasks boundaries are unknown, or when such boundaries are not well-defined (e.g., the data distribution continuously changes in a non-stationary manner).
For example, in image classification tasks in real-world scenarios, input images may exhibit several gradual changes through time, such as the zoom, illumination, or the angle of objects in the image. In this paper, we aim to reduce catastrophic forgetting in such difficult and relevant task agnostic cases.

It is long known that estimating the underlying posterior distribution can 
help mitigate catastrophic forgetting in neural networks~\citep{mccloskey1989catastrophic}.
Lately, it was found~\citep{kirkpatrick2017overcoming} that one can use the posterior to find confidence levels for weights, which in turn can be used to affect the weight plasticity. That is to say, weights with a lower confidence value may be changed more in subsequent tasks (i.e., since they are "less important" for previous tasks). This allows a natural transition between learned tasks while reducing the ill effect of catastrophic forgetting. Later, \cite{nguyen2017variational} also used variational Bayes to prevent catastrophic forgetting when the tasks arrive sequentially. In these works, each time a task switches, a new prior (or a regularization term) is added, which restricts the change of weights with high confidence (i.e., since they are "important" for previous tasks). However, in task agnostic scenarios this approach cannot be used, since the task boundary is unknown, or does not exist. 

To prevent catastrophic forgetting in task agnostic scenarios, we propose using the online version of variational Bayes, which updates the approximate posterior using only the current data sample without any knowledge on task switches. For the method to be effective, we aim to approximate the Bayes update as accurately as possible. Therefore, we derive novel fixed point equations for the online variational Bayes optimization problem under Gaussian parametric distributions (with either full,  matrix variate, or diagonal covariance). Based on our theoretical results, for each Gaussian distribution, we propose the \textit{Fixed-point Operator for Online Variational Bayes} (FOO-VB) algorithm.
Our experiments demonstrate a significant improvement over existing algorithms in task agnostic continual learning scenarios. 

\section{Related Work}
\paragraph{Bayesian neural networks}
Bayesian inference for neural networks has been a subject of interest over many years. As exact Bayesian inference is intractable (for any realistic network size), much research has focused on approximation techniques. Most modern techniques stem from previous seminal works that used either a Laplace approximation~\citep{Mackay1992}, variational methods~\citep{Hinton1993}, or Monte Carlo methods~\citep{Neal94}. In recent years, many methods for approximating the posterior distribution have been suggested, falling into one of these categories. Those methods include assumed density filtering~\citep{soudry2014expectation,hernandez2015probabilistic}, approximate power Expectation Propagation~\citep{hernandez2016black}, Stochastic Langevin Gradient Descent~\citep{welling2011bayesian,balan2015bayesian}, incremental moment matching~\citep{Lee2017}, and variational Bayes~\citep{graves2011practical,blundell2015weight}. 

In this work, we focus on a variational Bayes approach. Practical variational Bayes for modern neural networks was first introduced by~\cite{graves2011practical}, where a parametric distribution was used to approximate the posterior distribution by minimizing the variational free energy. Calculating the variational free energy is intractable for general neural networks. Thus, \cite{graves2011practical}
estimated the gradients using a biased Monte Carlo method, and used stochastic gradient descent (SGD) to perform minimization.
In a later work,~\cite{blundell2015weight} used a re-parameterization trick to yield an unbiased estimator for the gradients. Variational Bayes methods were also used extensively on various probabilistic models including recurrent neural networks~\citep{graves2011practical}, auto-encoders~\citep{kingma2013auto}, and fully connected networks~\citep{blundell2015weight}. \cite{martens2015optimizing,zhang2017noisy,khan2018fast} suggested using the connection between natural gradient descent~\citep{amari1998natural} and
variational inference to perform natural gradient optimization in deep neural networks.

\paragraph{Fixed-point equations for the variation Bayes} In this work, we derive a novel fixed-point equation for the online variational Bayes optimization problem under Gaussian parametric distributions (with either full, matrix variate, or diagonal covariance) in the case of Bayesian neural networks. Fixed point equation for the variational Bayes optimization problem has been used in the cases of linear models \citep{NIPS2011_5c936263,NIPS2013_f4be0027} and Gaussian processes \citep{opper2009variational}. Those derivations can not be used in the case of Bayesian neural networks. For instance, in the case of Gaussian processes, the negative log-likelihood of a single data point is a function of only one random variable (the parameter of the kernel function). So, the non-diagonal elements of the posterior precision matrix are equal to those of the prior precision matrix, resulting in a closed-form solution to the posterior distribution. However, in the case of Bayesian neural networks, the negative log-likelihood of a single data point is a function of a random vector (the parameters vector). 
Another related work is \cite{kurle2019continual}, which suggested an algorithm for online inference for non-stationary streaming data. \cite{kurle2019continual} derived the first-order necessary conditions for the online variational Bayes optimization problem in the case of Bayesian neural networks and multivariate Gaussian distribution. \cite{kurle2019continual} used the first-order necessary conditions alongside a running memory to sequentially update the posterior parameters --- but only in the case of a diagonal Gaussian distribution. However, as we will show here, using non-diagonal covariance can significantly improve performance.

\paragraph{Continual learning}
Approaches for continual learning can be generally divided into four main categories: (1) Architectural approaches; (2) Rehearsal approaches; (3) Regularization approaches, and (4) Bayesian approaches. 

\textit{Architectural approaches} alter the network architecture to adapt to new tasks (e.g. \cite{rusu2016progressive,nagabandi2018deep}).
\textit{Rehearsal approaches} use external memory\footnote{\normalsize Rehearsal approaches may include GANs, which can be properly used as memory.} to allow re-training on stored examples from previous tasks (e.g. \cite{shin2017continual}).
\textit{Regularization approaches} use some penalty on deviations from previous task weights.
\textit{Bayesian approaches} use Bayes' rule with the previous task posterior as the current prior.
Each approach has pros and cons. For example, architectural approaches may result in extremely large architectures after multiple task switches. Rehearsal approaches may be problematic due to memory limitations or data availability (e.g. data from previous tasks may not be stored with the model due to privacy or proprietary reasons).
For an extensive review of continual learning methods see \cite{parisi2018continual}. Our paper, however, focuses on \textit{regularization} and \textit{Bayesian approaches} where the architecture is fixed, and no external memory is used to retrain on data from previous tasks.

In \emph{regularization} approaches, a regularization term is added to the loss function. \textit{Elastic weight consolidation} (EWC) proposed by~\cite{kirkpatrick2017overcoming}, slows changes in parameters that are important to the previous tasks by penalizing the difference between the previous task's optimal parameters and the current parameters. The importance of each parameter is measured using the diagonal of the Fisher information matrix.
\textit{Synaptic Intelligence} (SI) proposed by~\cite{zenke2017continual} also uses a penalty term, 
however, the importance is measured by the path length of updates on the previous task. \cite{chaudhry2018riemannian} propose an online generalization of EWC and SI to achieve better performance.
\textit{Progress \& compress} proposed by \cite{schwarz2018progress} uses a network with two components (a knowledge base and an active column) with EWC for continual learning in reinforcement learning. 
\textit{Memory Aware Synapses} (MAS) \citep{aljundi2018memory} also uses a penalty term, but measures weight importance by the sensitivity of the output function.
\textit{Learning without Forgetting} (LwF) proposed by~\cite{li2017learning} uses knowledge distillation to enforce the network outputs of the new task to be similar to the network outputs of previous tasks.

The \textit{Bayesian approaches} provide a solution to continual learning in the form of Bayes' rule. As data arrive sequentially, the posterior distribution of the parameters for the previous task is used as a prior for the new task. \textit{Variational Continual Learning} (VCL) proposed by~\cite{nguyen2017variational} uses online variational inference combined with the standard variational Bayes approach, \textit{Bayes By Backprop} (BBB) by~\cite{blundell2015weight}. They reduce catastrophic forgetting by replacing the prior on each task switch. In the BBB approach, a \textit{mean-field approximation} is applied, assuming weights are independent of each other (i.e. the covariance matrix is diagonal).
\cite{ritter2018online} suggests using Bayesian online learning with a Kronecker factored Laplace approximation to attain a non-diagonal method for reducing catastrophic forgetting. This allows the algorithm to exploit interactions between weights within the same layer.

\paragraph{Task-agnostic continual learning}
Most of the aforementioned methods, assume relaxed conditions, where tasks arrive sequentially, and the data distribution changes only on task switches.
Few previous works deal with task agnostic scenarios (when task boundaries are unknown or not defined), however, they all use the rehearsal approach \citep{rao2019continual,aljundi2019gradient,achille2018life}. We present an algorithm for task agnostic continual learning scenarios using a fixed architecture and without using external memory. Our work is orthogonal to those rehearsal approaches, and can potentially be combined with them.

\section{General Theoretical Background}
\label{subsec:Theory_background}
Bayesian inference \citep{gelman2013bayesian,bishop1995neural} requires a joint probability distribution over the target set $\bv D$ and the model
parameters $\bs \theta$ given the input set $\bv X$. This distribution can be written as
\begin{equation}
p\left(\bv D,\bs \theta| \bv X\right)=p\left(\bv D|\bs \theta, \bv X\right)p\left(\bs \theta| \bv X\right)\,,
\end{equation}
where $\bv X = \begin{bmatrix} \bv x_1 & \bv x_2 & \cdots & \bv x_N\end{bmatrix}^{\top}$ is the input set,
$\bv D = \begin{bmatrix} \bv y_1 & \bv y_2 & \cdots & \bv y_N\end{bmatrix}^{\top}$ is the target set, and $\bs \theta$ is the model parameters vector.
$p\left(\bv D| \bs \theta, \bv X\right)$ is the likelihood function of the target
set $\bv D$, and $p\left(\bs \theta| \bv X\right)$ is the prior distribution of
the model parameters $\bs \theta$. The posterior distribution of the model
parameters can be calculated
using Bayes' rule
\begin{equation}
p\left(\bs \theta|\bv D, \bv X\right)=\frac{p\left(\bv D|\bs \theta, \bv X\right)p\left(\bs \theta| \bv X\right)}{p\left(\bv D| \bv X\right)} \,, 
\end{equation}
where $p\left(\bv D| \bv X\right)$ is calculated using the sum rule. To simplify the notations we omit the conditioning on $\bv X$ for the remainder of this paper.

We focus on the online version of Bayesian inference, in which the data arrive sequentially, and we update the posterior distribution whenever new data arrive. In each step, the previous posterior distribution is used as the new prior distribution. Therefore, according to Bayes' rule, the posterior distribution at time $n$ is given by
\begin{equation}
p\left(\bs \theta|\bv D_{n}\right)=\frac{p\left(\bv D_{n}|\bs \theta\right)p\left(\bs \theta|\bv D_1,\cdots,\bv D_{n-1}\right)}{p\left(\bv D_{n}\right)} \,, \label{eq:online Bayes rule}
\end{equation}
Unfortunately, calculating the posterior distribution is intractable
for most practical probability models, and especially when using deep neural networks. Therefore, we will use variational methods to approximate the true posterior.

\paragraph{Variational Bayes}
In variational Bayes \citep{graves2011practical}, a parametric distribution $q\left(\bs \theta|\phi\right)$
is used for approximating the true posterior distribution $p\left(\bs \theta\right|\bv D)$
by (indirectly) minimizing the Kullback-Leibler (KL) divergence with the true posterior
distribution
\begin{equation}
\kl\left(q\left(\bs \theta|\phi\right)||p\left(\bs \theta|\bv D\right)\right)=
\mathbb{E}_{\bs \theta\sim q\left(\bs \theta|\phi\right)}\left[\log\frac{q\left(\bs \theta|\phi\right)}{p\left(\bs \theta|\bv D\right)}\right] \,. \label{eq:variational free energy}
\end{equation}
The optimal variational parameters ($\phi$) are the solution of the following optimization problem:
\begin{align}
& \arg\min_{\phi}\int q\left(\bs \theta|\phi\right)\log\frac{q\left(\bs \theta|\phi\right)}{p\left(\bs \theta|\bv D\right)}d\bs \theta 
=\arg\min_{\phi}\int q\left(\bs \theta|\phi\right)\log\frac{q\left(\bs \theta|\phi\right)}{p\left(\bv D|\bs \theta\right)p\left(\bs \theta\right)}d\bs \theta\nonumber \\
 & =\arg\min_{\phi}\mathbb{E}_{\bs \theta\sim q\left(\bs \theta|\phi\right)}\left[\log\left(q\left(\bs \theta|\phi\right)\right)\!-\!\log\left(p\left(\bs \theta\right)\right) \!+\!\mathrm{L}\left(\bs \theta\right)\right]\!, \label{eq:variational bayes}
\end{align}
where $\mathrm{L}\left(\bs \theta\right)=-\log \left(p\left(\bv D|\bs \theta\right)\right)$
is the log-likelihood cost function.\footnote{\normalsize Note that we define a cumulative log-likelihood cost function over the data.}
The KL divergence between the parametric distribution (approximate posterior) and the true posterior
distribution~\eqref{eq:variational free energy} is also known as the variational free energy.

In online variational Bayes \citep{broderick2013streaming}, one aims to find the posterior in an online
setting, where data arrive sequentially. Similar to Bayesian
inference, we use the previous approximated posterior as the new prior
distribution, and the optimization problem becomes:
\begin{equation}
\!\arg\!\min_{\phi}\mathbb{E}_{\bs \theta\sim q_{n}\left(\bs \theta|\phi\right)}\!\left[\log\left(q_{n}\!\left(\bs \theta|\phi\right)\right)\!-\!\log\left(q_{n-1}\!\left(\bs \theta\right)\right)\! +\!\mathrm{L}_n\!\left(\bs \theta\right) \right]\!.\label{eq:online variational bayes} 
\end{equation}

\section{Proposed Theoretical Approach}
\label{sec:theory proposed approach}
We present a method to mitigate catastrophic forgetting in task agnostic continual learning. We aim to approximate the intractable (exact) online Bayes update rule \eqref{eq:online Bayes rule} using the online variational Bayes optimization problem \eqref{eq:online variational bayes}. Therefore, we use a new prior for each mini-batch (as in online variational Bayes) instead of using one prior for all the data (as in variational Bayes). When a Gaussian distribution is used as the parametric distribution  $q\left(\bs \theta\right|\phi)$, one can find the fixed-point equations for the online variational Bayes optimization problem~\eqref{eq:online variational bayes}, i.e. the first-order necessary conditions. The fixed-point equations define the relation between the prior parameters and the posterior parameters.
Using the fixed-point equations we derive algorithms for task agnostic continual learning (see section \ref{sec:Algorithms}).

In the subsections to come, we derive novel fixed-point equations for the online variational Bayes~\eqref{eq:online variational bayes} for multivariate Gaussian, Matrix variate Gaussian, and diagonal Gaussian distributions.

\subsection{Fixed-point equations for  multivariate Gaussian}
\label{subsec:Gaussian approximation}
In this subsection we focus on our most general case, in which the parametric distribution $q_n\left(\bs \theta|\phi\right)$ and
the prior distribution $q_{n-1}\left(\bs \theta\right)$ are multivariate Gaussian. Namely,
\begin{align}
q_{n}\left(\bs \theta|\phi\right)&=\mathcal{N}\left(\bs \theta|\bs \mu,\bs \Sigma\right), ~ \, ~ q_{n-1}\left(\bs \theta\right)=\mathcal{N}\left(\bs \theta|\bv m,\bv V\right).
\end{align}
To find the fixed-point equations of the optimization problem in~\eqref{eq:online variational bayes} in the case of a Gaussian distribution,
we define the following deterministic transformation:
\begin{align}
\bs \theta =\bs \mu+\bv A\bs \epsilon \,,\label{eq:gaussain deterministic transformation}
\end{align}
where $\phi =\left(\bs \mu,\bs \Sigma\right),\quad \bs \Sigma = \bv A\bv A^\top, \quad \bs \epsilon \sim\mathcal{N}\left(0,\mathbf{I}\right)$.

Using the first-order necessary conditions on~\eqref{eq:online variational bayes} for the optimal $\bs \mu$ and $\bv A$ (see Appendix \ref{app: Derivation of Gaussian first order necessary conditions} for details) we obtain the following equations
\begin{align}
\bs \mu = \bv m-\bv V\mathbb{E}_{\bs \epsilon}\left[\nabla\mathrm{L}_n\left(\bs \theta\right)\right], \quad
\bv A\bv A^{\top}+\bv V\mathbb{E}_{\bs \epsilon}\left[\nabla\mathrm{L}_n\left(\bs \theta\right)\bs \epsilon^{\top}\right]\bv A^{\top}-\bv V=0 \,. \label{eq:Gaussian first order necessary conditions}
\end{align}
In Lemma \ref{lemma}, we characterize the full set of solutions of the above quadratic equation (the proof can be found in Appendix \ref{app:Proof of Lemma}). 
\begin{lemma} \label{lemma}
Let $\bv T\in \mathbb{R}^{N \times N},\bv M \in \mathbb{R}^{N \times N},\bv M=\bv M^{\top}, \bv M \succ 0, \bv X \in \mathbb{R}^{N \times N}$. 
The full set of solutions of 
\begin{equation}
\bv X\bv X^{\top}+ \bv M \bv T\bv X^{\top}-\bv M=0 \label{eq:Matrix equation}
\end{equation}
is given by
\begin{equation*}
\bv X = \bv D \bv Q - \frac{1}{2}\bv M \bv T\,,
\end{equation*}
where 
\begin{align*}
\bv B = \bv M + \frac{1}{4}\bv M \bv T \bv T^{\top} \bv M, \quad
\bv D = \bv B^{1/2}, \quad
\end{align*}
and $\bv Q \in \mathbb{R}^{N \times N}$ is an orthogonal matrix such that $\bv D \bv Q \bv T^{\top} \bv M$ is a symmetric matrix.
\end{lemma}
In Corollary \ref{corollary: full set of solutions} we demonstrate that $\bs \Sigma$ has a set of multiple solutions.
\begin{corollary} \label{corollary: full set of solutions}
The full set of solutions for the optimal covariance matrix of the posterior distribution is given by
\begin{align}
    \bs \Sigma &= \bv V + \frac{1}{2}\bv V \mathbb{E}_{\bs \epsilon}\left[\nabla\mathrm{L}_n\left(\bs \theta\right)\bs \epsilon^{\top}\right]\mathbb{E}_{\bs \epsilon}\left[\nabla\mathrm{L}_n\left(\bs \theta\right)\bs \epsilon^{\top}\right]^{\top}\bv V \nonumber \\
    &\quad -\frac{1}{2}\left(\bv D\bv Q\mathbb{E}_{\bs \epsilon}\left[\nabla\mathrm{L}_n\left(\bs \theta\right)\bs \epsilon^{\top}\right]^{\top} \bv V + \bv V \mathbb{E}_{\bs \epsilon}\left[\nabla\mathrm{L}_n\left(\bs \theta\right)\bs \epsilon^{\top}\right] \bv Q^{\top} \bv D^{\top} \right)
\end{align}
where 
\begin{align}
    \bv D = \left(\bv V + \frac{1}{4}\bv V \mathbb{E}_{\bs \epsilon}\left[\nabla\mathrm{L}_n\left(\bs \theta\right)\bs \epsilon^{\top}\right]\mathbb{E}_{\bs \epsilon}\left[\nabla\mathrm{L}_n\left(\bs \theta\right)\bs \epsilon^{\top}\right]^{\top}\bv V \right)^{1/2}
\end{align}
and $\bv Q$ is an orthogonal matrix such that $\bv D \bv Q \mathbb{E}_{\bs \epsilon}\left[\nabla\mathrm{L}_n\left(\bs \theta\right)\bs \epsilon^{\top}\right]^{\top}\bv V$ is a symmetric matrix.
\end{corollary}
Next, we characterize a single solution for the quadratic equation using the following Lemma (the proof can be found in Appendix \ref{app:Proof of Lemma1}):
\begin{lemma} \label{lemma1}
In this Lemma we use the notations of Lemma \ref{lemma}. Let $\bv Q = \bv S \bv W^{\top}$ such that $\bv S, \bv W$ are the left and right singular matrices of the Singular Value Decomposition (SVD) of $\bv D^{-1} \bv M \bv T$. Then $\bv Q$ is an orthogonal matrix and $\bv D \bv Q \bv T^{\top} \bv M$ is a symmetric matrix.
\end{lemma}
Lemma \ref{lemma} and \ref{lemma1} together reveal the fixed-point equations (we substitute $\bv M= \bv V$ and $\bv T = \mathbb{E}_{\bs \epsilon}\left[\nabla\mathrm{L}_n\left(\bs \theta\right)\bs \epsilon^{\top}\right]$)
\begin{align}
\bs \mu = \bv m-\bv V\mathbb{E}_{\bs \epsilon}\left[\nabla\mathrm{L}_n\left(\bs \theta\right)\right], \quad
\bv A = \bv D \bv Q - \frac{1}{2}\bv V \mathbb{E}_{\bs \epsilon}\left[\nabla\mathrm{L}_n\left(\bs \theta\right)\bs \epsilon^{\top}\right] \,. \label{eq:multivariate Gaussian implicit solution}
\end{align}
In case $\bv B$ is not invertible (or, more often, has a large condition number), the fixed-point equations can be derived using Lemma \ref{lemma2}, which can be found in Appendix \ref{app:lemma2}.

\subsection{Fixed-point equations for matrix variate Gaussian}
\label{sub:Matrix variate Gaussian approximation}
We now focus on the case in which the parametric distribution $q_n\left(\bs \theta|\phi\right)$ and
the prior distribution $q_{n-1}\left(\bs \theta\right)$ are multivariate Gaussian whose covariance matrix is a Kronecker product of two PD matrices \citep{gupta2018matrix}. This type of distribution is also known as Kronecker-factored Gaussian. Therefore:
\begin{align}
q_{n}\left(\bs \theta|\phi\right)=\mathcal{N}\left(\bs \theta|\bs \mu,\bs \Sigma_1\otimes\bs \Sigma_2\right), \quad  q_{n-1}\left(\bs \theta\right)=\mathcal{N}\left(\bs \theta|\bv m,\bv V_1\otimes\bv  V_2\right)\,,
\end{align}
where $\bv V_1,\bs \Sigma_1\in\mathbb{R}^{d_1\times d_1}$ (variance among-column) and $\bv V_2, \bs \Sigma_2\in\mathbb{R}^{d_2\times d_2}$ (variance among-row).
To find the fixed-point equations of the optimization problem in \eqref{eq:online variational bayes} in the case of a matrix variate Gaussian distribution, we define a deterministic transformation
\begin{align}
\bs \theta =\bs \mu+\left(\bv A\otimes\bv B\right)\bs \epsilon
\,, \label{eq:Kronecker-factored deterministic transformation}
\end{align}
where the distribution parameters are $\phi =\left(\bs \mu,\bs \Sigma_1, \bs \Sigma_2\right)$ and
$\bs \Sigma_1 = \bv A\bv A^\top, \bs \Sigma_2 = \bv B\bv B^\top, \bs \epsilon \sim\mathcal{N}\left(0,\mathbf{I}\right)$.

We use the first-order necessary conditions for the optimal $\bs \mu, \bv A$, and $\bv B$ (see Appendix \ref{app:Derivation of Kronecker-factored necessary conditions} for additional details)
\begin{align}
&\bs \mu = \bv m-\left(\bv V_1\otimes \bv V_2\right)\mathbb{E}_{\bs \epsilon}\left[\nabla\mathrm{L}_n\left(\bs \theta\right)\right]  \label{eq:Kronecker-factored necessary conditions} \\
&\trace\left(\bv V_2^{-1}\bs \Sigma_2\right)\bv A\bv A^{\top}+\bv V_1\mathbb{E}_{\bs \epsilon}\left[\Psi^{\top}\bv B\Phi\right]\bv A^{\top}-p\bv V_1=0 \nonumber \\
&\trace\left(\bv V_1^{-1}\bs \Sigma_1\right)\bv B\bv B^{\top}+\bv V_2\mathbb{E}_{\bs \epsilon}\left[\Psi\bv A\Phi^{\top}\right]\bv B^{\top}-n\bv V_2=0 
\,,\nonumber
\end{align}
where $\Psi,\Phi\in\mathbb{R}^{p\times n}$ such that $\Vectorization(\Psi)=\nabla\mathrm{L}_n\left(\bs \theta\right)$ and $\Vectorization(\Phi)=\bs \epsilon$. In Appendix \ref{app:fixed-point equations for the  matrix variate Gaussian} we use Lemma \ref{lemma1} to derive the fixed-point equations for \eqref{eq:Kronecker-factored necessary conditions}.

\subsection{Fixed-point equations for diagonal Gaussian}
\label{sub:Diagonal Gaussian approximation}
Now consider a parametric distribution $q\left(\bs \theta|\phi\right)$ and
a prior distribution that are both Gaussian with a diagonal covariance matrix (i.e. mean-field approximation). Therefore
\begin{align}
q_{n}\left(\bs \theta|\phi\right) =\prod_{i}\mathcal{N}\left(\theta_{i}|\mu_{i},\sigma_{i}^{2}\right), \quad q_{n-1}\left(\bs \theta\right)=\prod_{i}\mathcal{N}\left(\theta_{i}|m_{i},v_{i}^{2}\right)\,. 
\end{align}
To derive the fixed-point equations of the optimization problem in \eqref{eq:online variational bayes}, we once more define a deterministic transformation
\begin{align}
\theta_{i} =\mu_{i}+\epsilon_{i}\sigma_{i} \,, \label{eq:mean-field deterministic transformation}
\end{align}
where $\phi =\left(\bs \mu,\bs \sigma\right), \quad \epsilon_{i} \sim\mathcal{N}\left(0,1\right)$.

We use the first-order necessary conditions for the optimal $\mu_i$ and $\sigma_i$
\begin{align}
\mu_{i}=m_{i}-v_{i}^{2}\mathbb{E}_{\epsilon}\left[\frac{\partial \mathrm{L}_n\left(\bs \theta\right)}{\partial\theta_{i}}\right], \quad
\sigma_{i}^{2}+\sigma_{i}v_{i}^{2}\mathbb{E}_{\epsilon}\left[\frac{\partial \mathrm{L}_n\left(\bs \theta\right)}{\partial\theta_{i}}\epsilon_{i}\right]-v_{i}^{2}=0\,.
\end{align}
Since this is a special case of \eqref{eq:Gaussian first order necessary conditions} (where the covariance matrix is diagonal), one can use \eqref{eq:multivariate Gaussian implicit solution} to derive the following fixed-point equations:
\begin{align}
\mu_{i}&=m_{i}-v_{i}^{2}\mathbb{E}_{\epsilon}\left[\frac{\partial \mathrm{L}_n\left(\bs \theta\right)}{\partial\theta_{i}}\right] \label{eq:diagonal implicit solution} \\
\sigma_{i}&=v_{i}\sqrt{\!1\!+\!\left(\!\frac{1}{2}v_{i}\mathbb{E}_{\epsilon}\left[\frac{\partial \mathrm{L}_n\left(\!\bs \theta\right)}{\partial\theta_{i}}\epsilon_{i}\right]\!\right)^{2}} \!\!\!-\!\frac{1}{2}v_{i}^{2}\mathbb{E}_{\epsilon}\!\left[\!\frac{\partial \mathrm{L}_n\left(\bs \theta\right)}{\partial\theta_{i}}\epsilon_{i}\right]\!. \nonumber 
\end{align}
\section{Proposed Algorithms}
\label{sec:Algorithms}
\subsection{From theory to practice}
The fixed-point equations we derived in section \ref{sec:theory proposed approach} to the optimization problem \eqref{eq:online variational bayes} are only implicit solutions.
Note for instance that the equations in \eqref{eq:multivariate Gaussian implicit solution} include the derivative $\nabla\mathrm{L}_n\left(\bs \theta\right)$, which is a function of $\phi$ (the unknown posterior parameters). 
One possible approach to find a solution for the fixed-point equations is to iterate them. In certain simple linear models in offline variational Bayes, a similar approach can be proven to converge  \citep{pmlr-v51-sheth16}. Here, since we are at an online setting, we take a single explicit iteration of the fixed-point equation for each mini-batch, i.e. we evaluate the derivative $\nabla\mathrm{L}_n\left(\bs \theta\right)$ using the prior parameters.
This is done during the multiple passes through the data, as in Assumed Density Filter (ADF) \citep{soudry2014expectation,hernandez2015probabilistic}.

The fixed-point equations iteration consists of an expectation term w.r.t. $\epsilon$. 
We use Monte Carlo samples to estimate those expectations.

\subsection{Algorithms}
\paragraph{FOO-VB}
Using the relaxations above, we present the \textit{Fixed-point operator for online variational Bayes} (FOO-VB)  algorithmic framework (Algorithm \ref{algo:Fixed-point operator for online variational Bayes}),
including three different specific variants: multivariate Gaussian, matrix variate Gaussian, and diagonal Gaussian (Algorithms \ref{algo:multivariate Gaussian}, \ref{algo:KFAC Gaussian},  and  \ref{algo:Diagonal Gaussian}; the last two variants can be found in Appendix \ref{app:algorithms}).

Algorithm \ref{algo:Fixed-point operator for online variational Bayes} describes the general framework in which we update the posterior distribution as we iterate over the data.
For each mini-batch we use $K$ Monte Carlo samples of the neural network weights using the current prior distribution $\phi_{n-1}$ (steps 1 \& 2), and then calculate the gradient w.r.t. these randomized weights (step 3). Lastly, we update the posterior parameters using the estimations and the update rules derived in section \ref{sec:theory proposed approach} (step 4).

\begin{algorithm}[h!]
\caption{Fixed-point operator for online variational Bayes (FOO-VB)}
\begin{description}
\item [{Initialize}] Prior parameters $\phi_0$, Number of iterations $N_{max}$,  \\ Number of Monte Carlo samples $K$
\item [{for}] 
\begin{description}
\item $n=1,\dots,N_{max}$ sample a mini-batch 
\item [{for}] $k = 1,\dots,K$
\begin{description}
\item Sample $\bs \epsilon^{(k)} \sim \mathcal{N}\left(0,\mathbf{I}\right)$  \hfill\algorithmiccomment{step 1}
\item $\bs \theta^{(k)} = \textsc{Transform} \left(\bs \epsilon^{(k)},\phi_{n-1}\right)$ \hfill\algorithmiccomment{step 2}
\item $\bv g^{(k)} = \nabla\mathrm{L}_n \left(\bs \theta^{(k)}\right)$ \hfill\algorithmiccomment{step 3}
\end{description}
\item [{end}]
\item $\phi_n \xleftarrow{} \textsc{Update}\left(\phi_{n-1},\bs \epsilon^{(1,\dots,K)},\bv g^{(1,\dots,K)} \right)$\, \hfill\algorithmiccomment{step 4}
\end{description}
\label{algo:Fixed-point operator for online variational Bayes}
\end{description}
\end{algorithm}

The deterministic transformation (step 2) and the posterior update rule (step 4) differ for each distribution. Algorithms \ref{algo:multivariate Gaussian}, \ref{algo:KFAC Gaussian} and \ref{algo:Diagonal Gaussian} described those steps for multivariate Gaussian,  matrix variate Gaussian, and diagonal Gaussian, respectively (Algorithms \ref{algo:KFAC Gaussian} and \ref{algo:Diagonal Gaussian} can be found in Appendix \ref{app:algorithms}).

\begin{algorithm}[h!]
\caption{Methods for the multivariate Gaussian version}
\begin{description}
\item [{\textsc{Transform}$\left(\bs \epsilon,\phi = \left(\bs \mu, \bv A\right)\right):$}] ~
\begin{description}
\item $\bs \theta =\bs \mu+\bv A\bs \epsilon$
\item $\bs \Sigma = \bv A\bv A^\top $
\end{description}
\item [{\textsc{Update}$\left(\phi_{n-1} = \left(\bs \mu,\bv A\right), \bs \epsilon^{(1..K)}, \bv g^{(1..K)}\right):$}] ~
\begin{description}
\item $\bar{\bv E}_{1} = \frac{1}{K}\sum_{k=1}^{K}\bv g^{(k)}$
\item $\bar{\bv E}_{2} = \frac{1}{K}\sum_{k=1}^{K}\bv g^{(k)}\left(\bs \epsilon^{(k)}\right)^{\top}$
\item $\bs \mu \xleftarrow{} \bs \mu -\bv A \bv A^{\top} \bar{\bv E}_{1}$
\item $\bv A \xleftarrow{} \bv X \text{ s.t. } \bv X\bv X^{\top}+ \bv A \bv A^{\top} \bar{\bv E}_{2} \bv X^{\top}-\bv A \bv A^{\top}=0 $
\item (This matrix equation is solved using Lemma \ref{lemma1})
\end{description}
\label{algo:multivariate Gaussian}
\end{description}
\end{algorithm}

\paragraph{Feasibility and complexity}
As Deep Neural Networks (DNN) often contain millions of parameters, it is infeasible to store the full covariance matrix between all parameters. A very common relaxation is to assume this distribution is factored between the layers (i.e. independent layers), so that the covariance matrix is a block diagonal matrix between layers.
Even with this relaxation, the multivariate Gaussian version of FOO-VB is impractical due to memory limitations. For example, storing the matrix $\bv A$ for a fully connected layer with $400$ inputs and $400$ outputs will require $\sim 95 \text{GB}$ (using 32-bit floating point).

We avoid working with such extremely large covariance matrices, by 
factoring them into a Kronecker product of two much smaller matrices (matrix variate Gaussian). Alternatively, we employ a diagonal covariance matrix (diagonal Gaussian).

The matrix variate Gaussian approximation enables us to store and apply mathematical operations on matrices of a practical size. For example, storing the matrices $\bv A$ and $\bv B$ of the matrix variate version for a fully connected layer with $400$ inputs and $400$ outputs will require $\sim 1.2 \text{MB}$.
In terms of runtime, this version requires four SVD decompositions for each layer in addition to the $K=2500$ Monte-Carlo (MC) samples for each iteration. To reduce the runtime, one can parallelize the SVD (as in \cite{BERRY1989191}) and the MC sampling. 

The diagonal approximation is our lightest version, with a memory footprint of only twice that of the regular network (to store the mean and variance of every weight). 
The $K=10$ MC samples are the only overhead over a standard SGD optimizer, and so the runtime is linear w.r.t. the number of MC samples. See Appendix \ref{appendix:complexity discussion} for computational complexity analysis.

\subsection{Theoretical properties of diagonal FOO-VB}
\label{theoretical_properties}
We present theoretical properties of FOO-VB. For simplicity, we focus on the Diagonal Gaussian version.
This version consists of a gradient descent algorithm
for $\mu$, and a recursive update rule for $\sigma$. The learning rate of $\mu_{i}$ is proportional to the uncertainty in the parameter $\theta_{i}$ according to the prior distribution. During the learning process, as more data is seen, the learning rate decreases for parameters with a high degree of certainty, and increases for parameters with a high degree of uncertainty. Next, we establish this intuitive idea more precisely.

It is easy to verify that the update rule for $\sigma$ is a strictly monotonically decreasing
function of $\mathbb{E}_{\epsilon}\left[\frac{\partial \mathrm{L}_n\left(\bs \theta\right)}{\partial\theta_{i}}\epsilon_{i}\right]$.
Therefore
\begin{align} 
\mathrm{E}_{\epsilon}\left[\frac{\partial L\left(\bs \theta\right)}{\partial\theta_{i}}\epsilon_{i}\right] & >0\Longrightarrow\sigma_{i}\left(n\right)<\sigma_{i}\left(n-1\right)\nonumber \\
\mathrm{E}_{\epsilon}\left[\frac{\partial L\left(\bs \theta\right)}{\partial\theta_{i}}\epsilon_{i}\right] & <0\Longrightarrow\sigma_{i}\left(n\right)>\sigma_{i}\left(n-1\right)\nonumber \\
\mathrm{E}_{\epsilon}\left[\frac{\partial L\left(\bs \theta\right)}{\partial\theta_{i}}\epsilon_{i}\right] & =0\Longrightarrow\sigma_{i}\left(n\right)=\sigma_{i}\left(n-1\right) \,.
\end{align}
Next, using a Taylor expansion for the loss, we show that for small values of $\sigma$, the quantity $\!\mathbb{E}_{\epsilon}\!\left[\!\frac{\partial \mathrm{L}_n\left(\bs \theta\right)}{\partial\theta_{i}}\epsilon_{i}\right]$ is equal to
\begin{align*}
\mathbb{E}_{\epsilon}\!\left[\left(\frac{\partial \mathrm{L}_n\left(\bv{\mu}\right)}{\partial\theta_{i}}+\sum_{j}\frac{\partial^{2}\mathrm{L}_n\left(\bv{\mu}\right)}{\partial\theta_{i}\partial\theta_{j}}\epsilon_{j}\sigma_{j}+O\left(\left\Vert \bv{\sigma}\right\Vert ^{2}\right)\right)\epsilon_{i}\right]\! 
=\frac{\partial^{2}\mathrm{L}_n\left(\bv{\mu}\right)}{\partial^{2}\theta_{i}}\sigma_{i}\!+\!O\left(\left\Vert \bv{\sigma}\right\Vert ^{2}\right)\,,
\end{align*}
where we used $\mathbb{E}_{\epsilon}\left[\epsilon_{i}\right]=0$ and $\mathbb{E}_{\epsilon}\left[\epsilon_{i}\epsilon_{j}\right]=\delta_{ij}$. Thus, in this case $\mathbb{E}_{\epsilon}\left[\frac{\partial \mathrm{L}_n\left(\bs \theta\right)}{\partial\theta_{i}}\epsilon_{i}\right]$ is a finite difference approximation to the component-wise product of the diagonal of the Hessian of the loss and the vector $\sigma$. Therefore, we expect the uncertainty (learning rate) to decrease in areas with positive curvature (e.g., near local minima), or increase in areas with high negative curvature (e.g., near maxima, or saddles). This seems like a ``sensible'' behavior of the algorithm, since we wish to converge to local minima, but escape saddles. This in contrast to many common optimization methods, which are either insensitive to the sign of the curvature, or use it the wrong way~\citep{Dauphin2014}.

In the case of a strongly convex loss, we prove a more rigorous statement in Appendix \ref{appendix:Proof of Theorem}.

\begin{theorem} \label{theorem}
Consider FOO-VB with a diagonal Gaussian distribution for $\bs \theta$. If ${\mathrm{L}_n}\left(\bs \theta\right)$ is a strongly convex function with parameter $m_{n}>0$ and a continuously
differentiable function over $\mathbb{R}^{n}$, then 
$\mathbf{\mathbb{E}}_{\epsilon}\left[\frac{\partial \mathrm{L}_n\left(\bs \theta\right)}{\partial\theta_{i}}\epsilon_{i}\right]\geq m_n\sigma_i>0$.
\end{theorem}

\begin{corollary} \label{corollary}
 If $\mathrm{L}_n\left(\bs \theta\right)$ is strongly convex (concave) for all $n\in \mathbb{N} $, then the sequence $\left\{\sigma_{i}(n)\right\}_{n=1}^{\infty}$
is strictly monotonically decreasing (increasing).
\end{corollary}

Furthermore, one can generalize these results and show that if a restriction of $\mathrm{L}_n\left(\bs \theta\right)$ to an axis $\theta_{i}$ is strongly convex (concave) for all $n\in \mathbb{N}$, then $\left\{\sigma_{i}(n)\right\}_{n=1}^{\infty}$
is monotonic decreasing (increasing).

Therefore, in the case of a strongly convex loss function, $\sigma_{i}=0$ in any stable point of \eqref{eq:diagonal implicit solution}, which means that we collapse to point estimation similar to SGD. However, for neural networks, $\sigma_{i}$ does not generally converge to zero. In this case, the stable point $\sigma_{i}=0$ is generally not unique, since $\mathbb{E}_{\epsilon}\left[\frac{\partial \mathrm{L}\left(\bs \theta\right)}{\partial\theta_{i}}\epsilon_{i}\right] $ implicitly depends on $\sigma_{i}$. 

In Appendix \ref{appendix:5000 epochs training} we show the histogram of STD (standard deviation) values on MNIST when training for 5000 epochs and demonstrate FOO-VB do not collapse to  point estimation.

\paragraph{FOO-VB in continual learning} In the case of over-parameterized models and continual learning, only a part of the weights is essential for each task. We hypothesize that if a weight $\theta_{i}$ is important to the current task, this implies that near the minimum, the function  $\mathrm{L}_i=\mathrm{L}(\bs \theta)|_{(\bs \theta)_i = \theta_i}$  is locally convex. Corollary \ref{corollary} suggests that in this case $\sigma_{i}$ would be small. In contrast, the loss will have a flat curvature in the direction of weights which are not important to the task. Therefore, these unimportant weights may have a large uncertainty $\sigma_{i}$.
Since FOO-VB in the Diagonal Gaussian version introduces the linkage between the learning rate and the uncertainty (STD), the training trajectories in the next task would be restricted along the less important weights leading to a good performance on the new task, while retaining the performance on the current task. The use of FOO-VB to continual learning exploits the inherent features of the algorithm, and it does not need any explicit information on tasks --- it is completely unaware of the notion of tasks. We show empirical evidence for our hypothesis in subsection \ref{subsec:Discrete Task-agnostic Permuted MNIST}

\section{Applications and Experiments}
\label{sec:Experiments}
\paragraph{Inapplicability of other CL algorithms}
In task-agnostic scenarios, previous optimization-based methods for continual learning are generally inapplicable, as they rely on taking some actions (e.g., changing parameters in the loss function) on task switch, which is undefined in those scenarios.
Nevertheless, one possible adaptation is to take the core action at every iteration instead of at every task switch. Doing so is impractical for many algorithms due to the computational complexity, but for a fair comparison we have succeeded to run both Online EWC \citep{chaudhry2018riemannian} and MAS \citep{aljundi2018memory} with such an adaptation. As for rehearsal approach, it is orthogonal to our approach and can be combined. However, we focus on the challenging real-world scenario in which we do not have access to any data from previous tasks. Thus, in our experiments we do not compare to rehearsal algorithms.

We experiment on a task agnostic variations of the Permuted MNIST benchmark for continual learning. The Permuted MNIST benchmark is a set of tasks constructed by a random permutation of MNIST pixels. Each task has a different permutation of pixels from the previous one. For all the experiments shown below, we conducted an extensive hyper-parameter search to find the best results for each algorithm. See additional details in Appendix \ref{appendix:Implementation details}. 

\subsection{Discrete task-agnostic Permuted MNIST}
\label{subsec:Discrete Task-agnostic Permuted MNIST}
\begin{figure}[ht]
\begin{center}
\includegraphics[width=0.7\columnwidth]{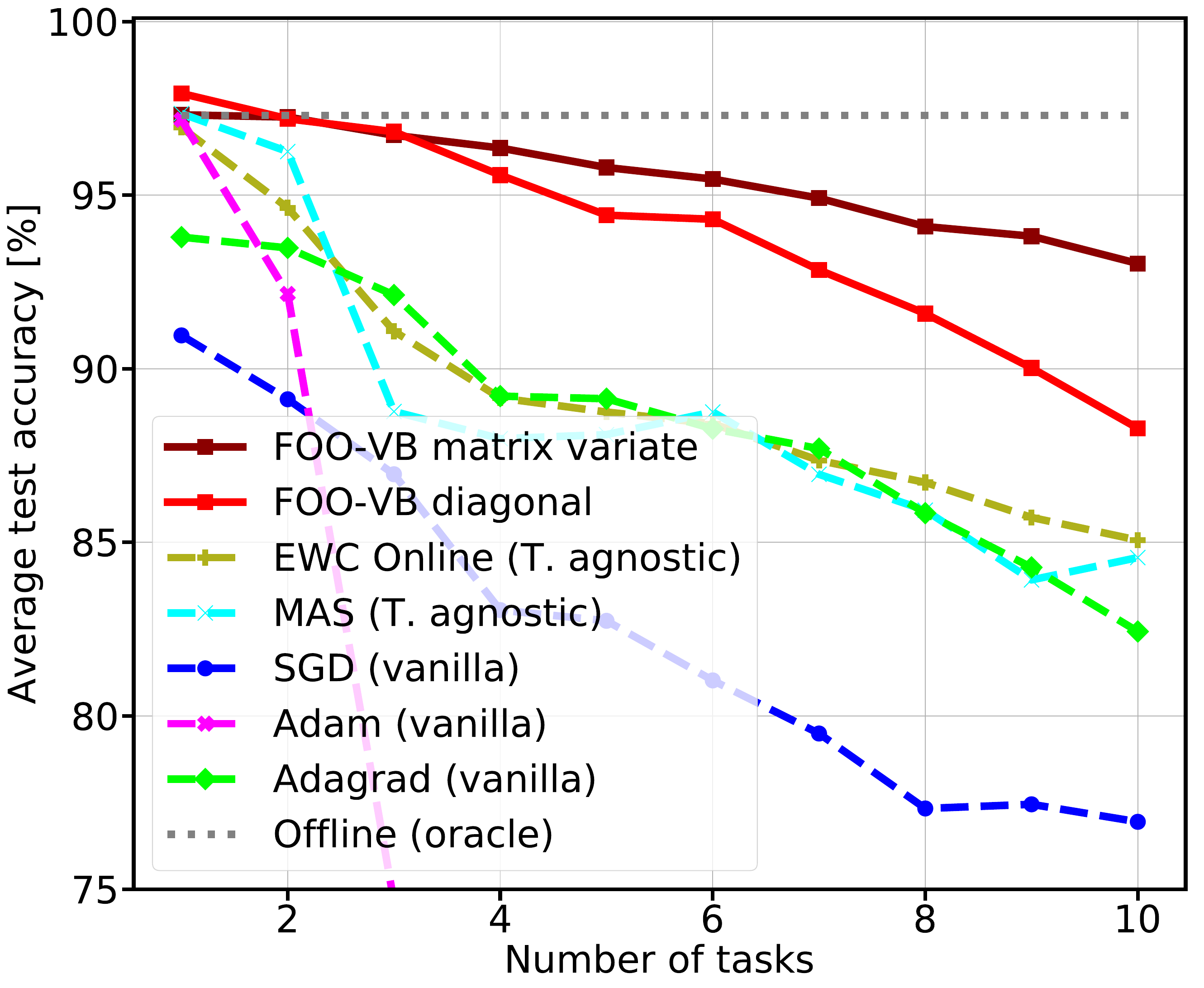}
\caption{\textbf{Discrete task-agnostic Permuted MNIST:} The average test accuracy on all seen tasks as a function of the number of tasks. The hyper-parameters of all algorithms were tuned to maximize the average accuracy over all 10 tasks (therefore some of the algorithms have a relatively low accuracy for the first task).
Offline (oracle) is a joint (i.e. not continual) training on all tasks.}
\label{fig:permuted_mnist_avg_acc}
\vspace{-1em}
\end{center}
\end{figure}

We evaluate the algorithms on a task-agnostic scenario where the task boundaries are unknown.
To do so, we use the Permuted MNIST benchmark for continual learning, but without informing the algorithms on task switches.
The network architecture is two hidden layers of width 100 (see additional details in Appendix \ref{appendix:Implementation details}).
In Figure \ref{fig:permuted_mnist_avg_acc}, we show the average test accuracy over all seen tasks as a function of the number of tasks.
Designed for task agnostic scenarios, our algorithms surpass all other task agnostic algorithms. The matrix variate Gaussian version of FOO-VB experiences only $\sim 2\%$ degradation in the average accuracy after 10 tasks. The diagonal Gaussian version of FOO-VB attains a good balance between high accuracy ($\sim 88\%$ after 10 tasks) and low computational complexity.

\begin{figure}[ht]
\begin{center}
\includegraphics[width=0.7\columnwidth]{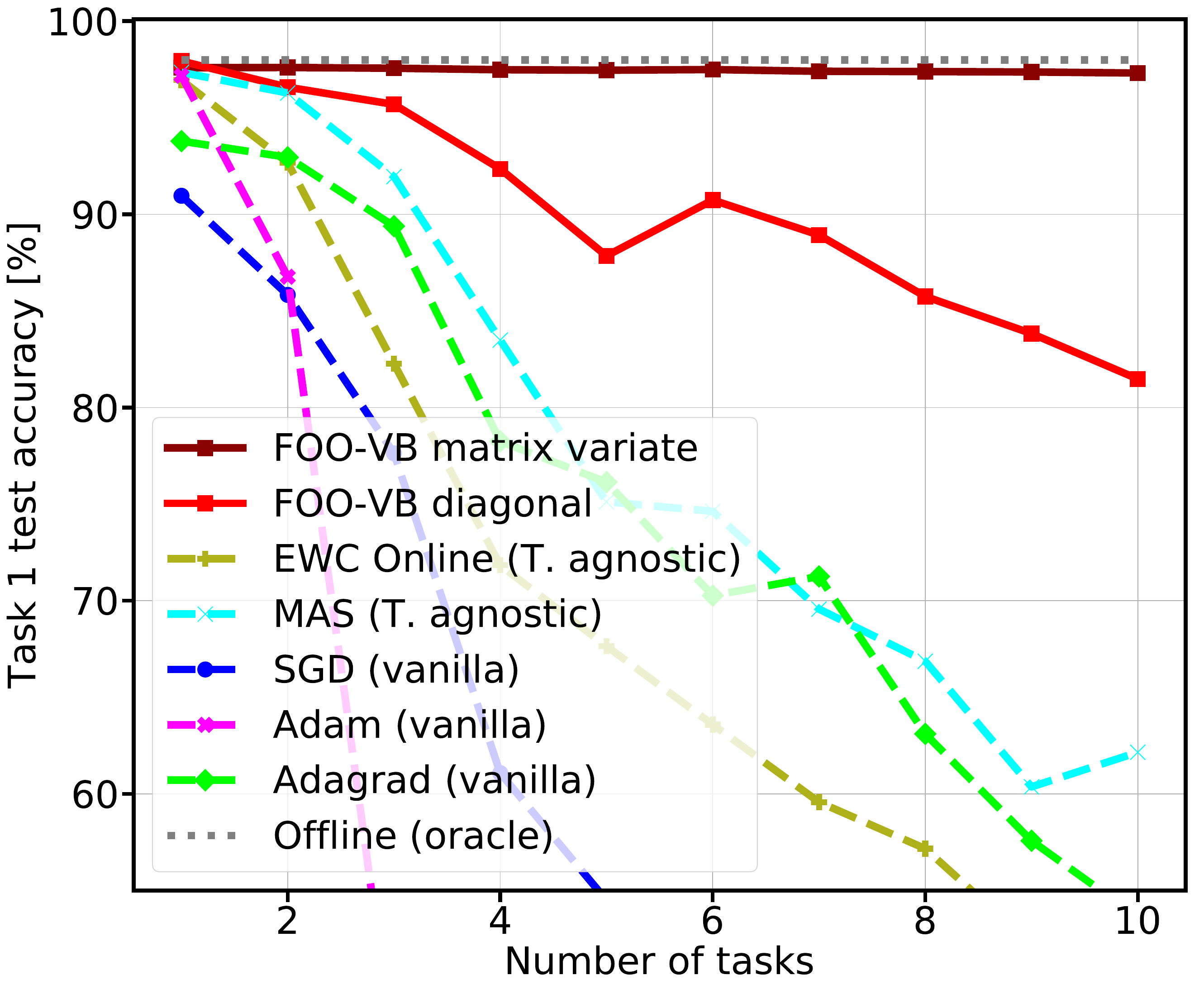}
\caption{Test accuracy on the first task for discrete task-agnostic permuted MNIST. Oracle is the accuracy of training on task \#1 only (i.e. not continual).}
\label{fig:permuted_mnist_first_task_acc}
\vspace{-1em}
\end{center}
\end{figure}

The average test accuracy over all tasks at the end of training implies a good balance between remembering previous tasks while adapting to new tasks. On the other hand, in Figure \ref{fig:permuted_mnist_first_task_acc} we show the test accuracy of the first task as a function of the number of seen tasks, which shows how well the algorithm remembers. The matrix variate Gaussian version of FOO-VB is being able to remember the first task almost perfectly (i.e. like the oracle), and the overall performance is limited only by the test accuracy of the current task. The diagonal Gaussian version of FOO-VB exhibits the next best performance, compared to other algorithms.

\begin{figure}[ht]
\begin{center}
\includegraphics[width=0.7\columnwidth]{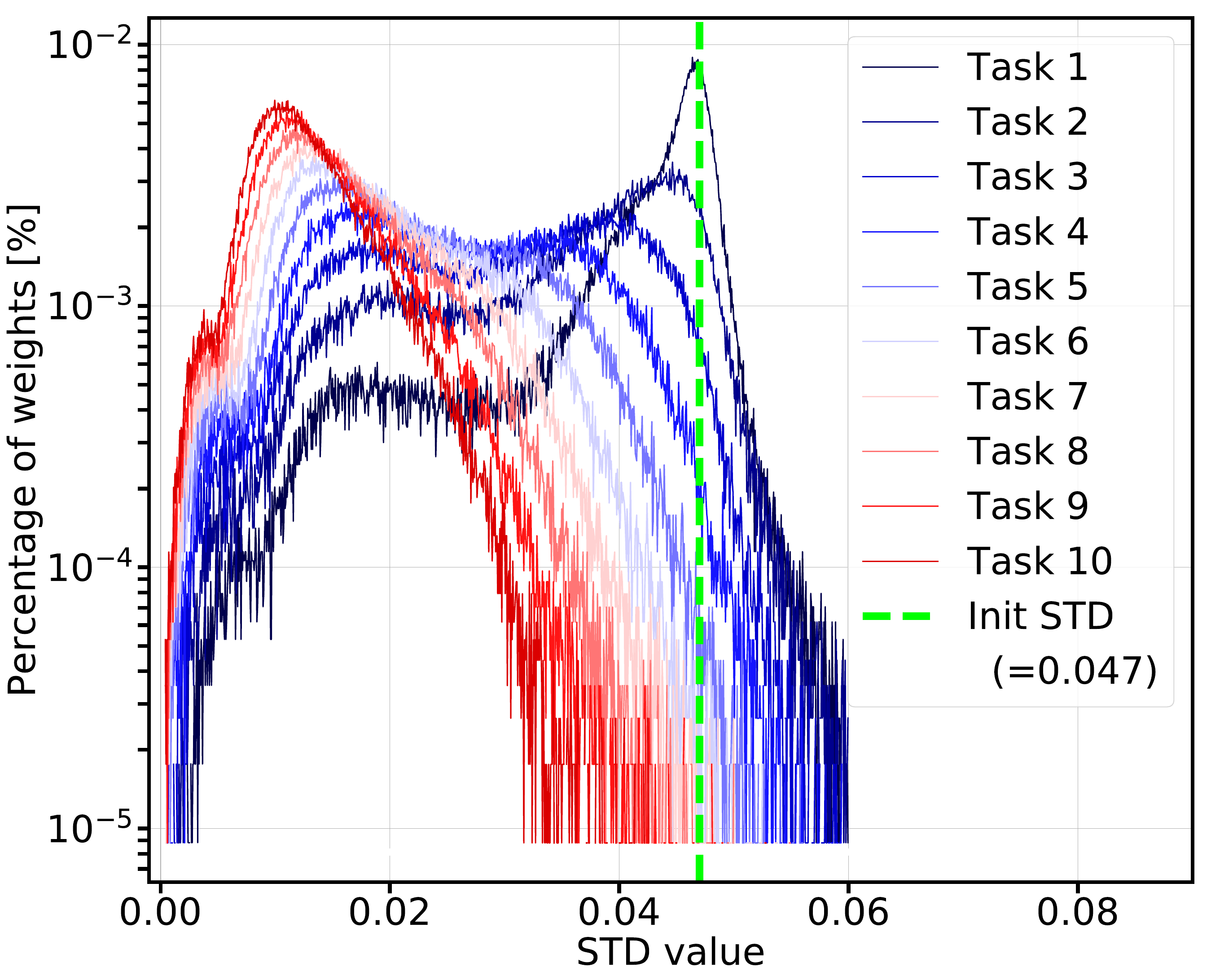}
\caption{The histogram of STD values at the end of the training process of each task. As training progresses, more weights are assigned with STD values lower than the initial value. The initial STD value is 0.047. Best seen in color.}
\label{fig:permuted_mnist_std_hist}
\vspace{-1em}
\end{center}
\end{figure}

We use the discrete task-agnostic Permuted MNIST experiment to examine our hypothesis of how the diagonal version of FOO-VB works in continual learning (subsection \ref{theoretical_properties}).
Figure \ref{fig:permuted_mnist_std_hist} shows the histogram of STD values at the end of the training process of each task.
The results show that after the first task, a large portion of the weights have STD values close to the initial value 0.047, while a small fraction of them have a much lower value. As training progresses, more weights are assigned with STD values much lower than the initial value.
These results support our hypothesis in subsection \ref{theoretical_properties} that only a small part of the weights is essential for each task, and as training progresses more wights have low STD values.

\subsection{Continuous task-agnostic Permuted MNIST}
\label{subsec:Continuous task-agnostic Permuted MNIST}
\begin{figure}[ht]
\begin{center}
\centerline{\includegraphics[width=0.7\columnwidth]{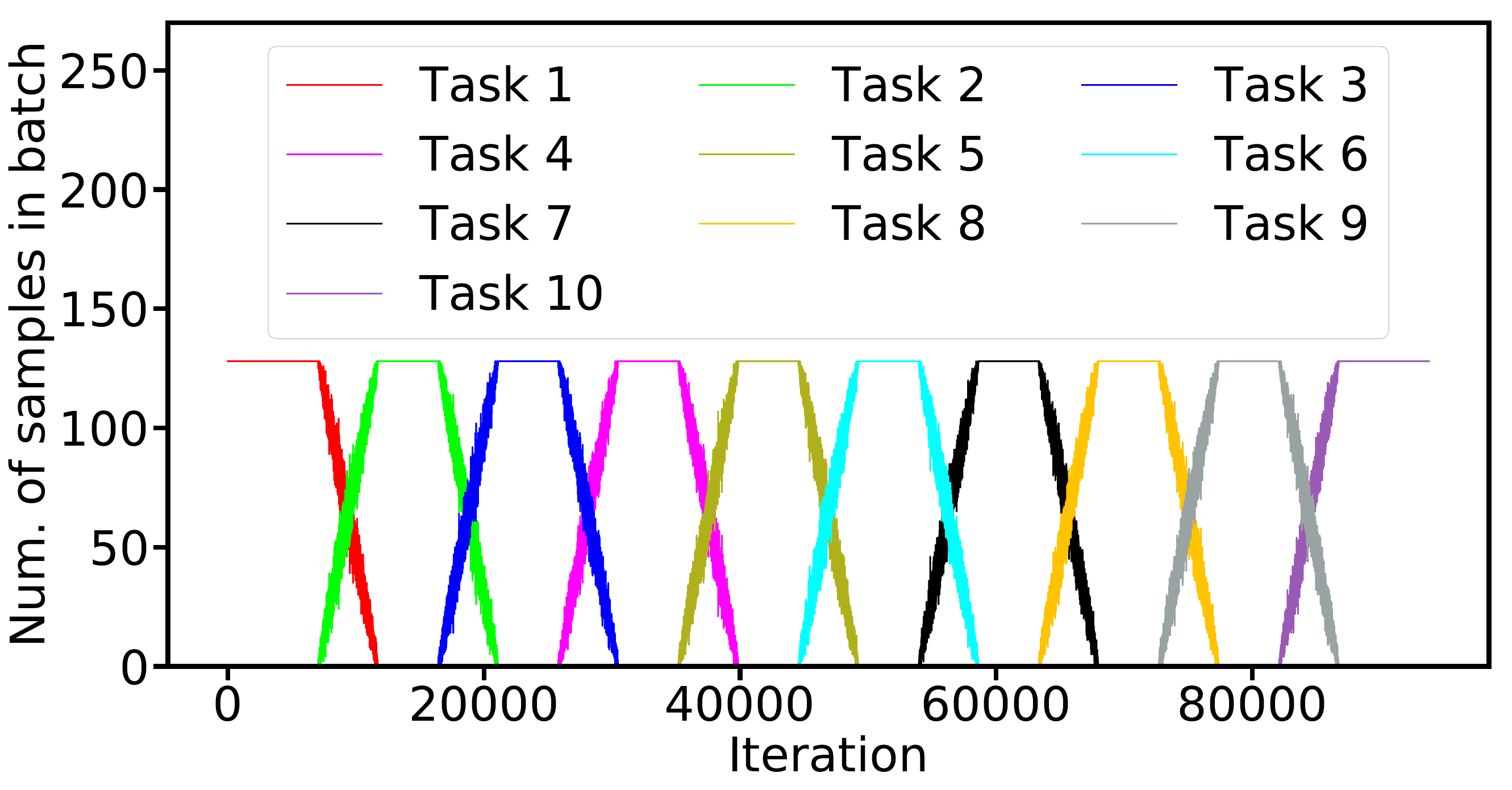}}
\caption{Distribution of samples from each task, as a function of iteration. The tasks are not changed abruptly but gradually -- i.e. task boundaries are undefined. Here, the number of samples from each task in each batch is a random variable drawn from this distribution, which changes with time (iterations).}
\label{fig:tasks_distribution}
\end{center}
\end{figure}

We consider the case where the transition between tasks occurs gradually over time, so the algorithm gets a mixture of samples from two \footnote{\normalsize The most challenging scenario is when mixing two different tasks. As we add more tasks, we are getting closer to offline (non CL) training.}
different tasks during the transition (Figure \ref{fig:tasks_distribution}) so the task boundaries are undefined.
In all task-agnostic scenarios, the algorithm does not have any knowledge of the distribution over the tasks.

The network architecture is two hidden layers of width 200 (see additional details in Appendix \ref{appendix:Implementation details}).
The output heads are shared among all tasks, task duration is $9380\times T$ iterations, where $T=10$ is the number of Tasks (corresponds to 20 epochs per task), and the algorithms are unaware to the number of tasks nor when the tasks are being switched.

\begin{figure}[ht]
\centering
\includegraphics[width=0.7\columnwidth]{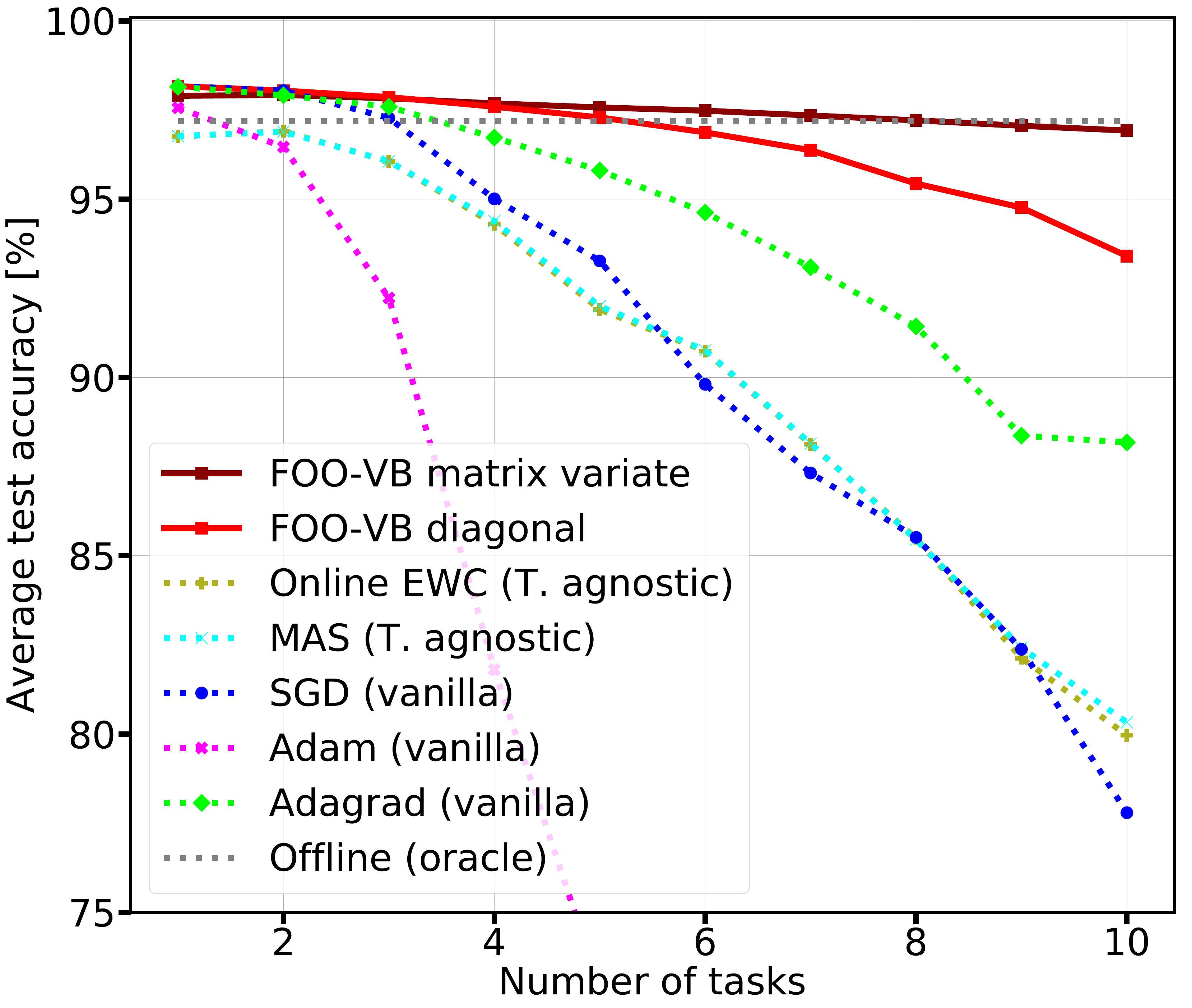}
\caption{\textbf{Continuous task-agnostic Permuted MNIST:} The average test accuracy on all seen tasks as a function of the number of tasks. Tasks are changing gradually over time as showed in Figure \ref{fig:tasks_distribution}. Offline (oracle) is a joint (not continual) training on all tasks.}
\label{fig:cont_permuted_mnist_results}
\end{figure}

The average test accuracy over all tasks for different numbers of tasks is presented in Figure \ref{fig:cont_permuted_mnist_results}.
As can be seen, the matrix variate Gaussian version of FOO-VB experiences less than 1\% degradation in the average accuracy after 10 tasks. Similarly to the discrete task-agnostic experiment (subsection \ref{subsec:Discrete Task-agnostic Permuted MNIST}), the diagonal Gaussian version of FOO-VB maintains a good balance between high accuracy ($\sim 94\%$ after 10 tasks) and low computational complexity.

\subsection{Task-aware continual learning on vision datasets}
We provide additional evaluation of FOO-VB using the experiment of  vision datasets conducted in \cite{ritter2018online}. This experiment is done in the task-aware scenario, and uses more complex datasets (such as CIFAR10 and SVHN) and architecture (LeNet). Thus, we use for this experiment the diagonal version of FOO-VB. The algorithms which FOO-VB is compared with are using the information on task-switch. Nevertheless, FOO-VB results are on par. See Appendix \ref{appendix:vision dataset} for the full details.

\section*{Conclusion}
In this work we aim to reduce catastrophic forgetting, in task agnostic scenarios (where task boundaries are unknown or not defined), using fixed architecture and without the use of external memory (i.e. without access to previous data, which can be restricted, e.g. due to privacy issues). This can allow deep neural networks to better adapt to new tasks without explicitly instructed to do so, enabling them to learn in real-world continual learning settings. 

Our method, FOO-VB, outperforms other continual learning methods in task agnostic scenarios. It relies on solid theoretical foundations, being derived from novel fixed-point equations of the online variational Bayes optimization problem. We derive two practical versions of the algorithm, to enable a trade-off between computational complexity vs. performance. 

There are many possible extensions and use cases for FOO-VB which were not explored in this work. One possible extension is to incorporate FOO-VB in the framework of Meta-Learning \citep{pmlr-v70-finn17a} to address more challenging scenarios. Another direction is to use FOO-VB to improve GAN stability \citep{thanhtung2018catastrophic}. During the training process, the Discriminator exhibit catastrophic forgetting, since the Generator output distribution changes gradually. FOO-VB fit this scenario naturally since it has no well defined task boundaries. Indeed, some of thus extensions were consider in \cite{he2019task} using the diagonal version of FOO-VB (published in our preliminary pre-print \citep{zeno2019task}).
Last, but not least, FOO-VB could be potentially useful in reinforcement learning, which often includes non-stationary environments.

\subsection*{Acknowledgments}
The authors are grateful to N. Merlis for insightful discussions and helpful comments on the manuscript. The research of DS was supported by the Israel Science Foundation (grant No. 1308/18), and by the Israel Inovation Authority (the Avatar Consortium).
\section*{Appendix}
\section{Derivation of \eqref{eq:Gaussian first order necessary conditions}}
\label{app: Derivation of Gaussian first order necessary conditions}
In this section, we provide additional details on the derivation of \eqref{eq:Gaussian first order necessary conditions}. The objective function is 
\begin{align}
f\left(\bs \mu,\bs \Sigma\right)&=\frac{1}{2}\left[\log\frac{\det{\left(\bv V\right)}}{\det{\left(\bs \Sigma\right)}}-N+\trace\left(\bv V^{-1}\bs \Sigma\right)+\left(\bv m-\bs \mu\right)^{\top}\bv V^{-1}\left(\bv m-\bs \mu\right)\right] +
\mathbb{E}_{\bs \theta}\left[\mathrm{L}_n\left(\bs \theta\right)\right]\,.
\end{align}
To solve the optimization problem in \eqref{eq:online variational bayes} in the case of Gaussian approximation, we use the deterministic transformation \eqref{eq:gaussain deterministic transformation}. 
To calculate the first derivative of the objective function we use the following identities:
\begin{align}
\mathbb{E}_{\bs \theta}\left[\mathrm{L}_n\left(\bs \theta\right)\right] &= 
\mathbb{E}_{\bs \epsilon}\left[\mathrm{L}_n\left(\bs \theta\right)\right] \\
\frac{\partial\mathbb{E}_{\bs \epsilon}\left[\mathrm{L}_n\left(\bs \theta\right)\right]}{\partial A_{i,j}}
&= \mathbb{E}_{\bs \epsilon}\left[\frac{\partial\mathrm{L}_n\left(\bs \theta\right)}{\partial\theta_i}\epsilon_j\right] \\ 
\frac{\partial\trace\left(\bv V^{-1}\bs \Sigma\right)}{\partial A_{i,j}} &= 2\sum_{n}V_{i,n}^{-1}A_{n,j} \\
\frac{\partial\log|\det{\left(\bv A\right)}|}{\partial A_{i,j}}&=A^{-\top}_{i,j} \,.
\end{align}
We use the first-order necessary conditions for the optimal $\bs \mu$:
\begin{equation}
-\bv V^{-1}\left(\bv m-\bs \mu\right)+\mathbb{E}_{\bs \epsilon}\left[\nabla\mathrm{L}_n\left(\bs \theta\right)\right] = 0\,.
\end{equation}
And so we obtained \eqref{eq:Gaussian first order necessary conditions}.

Next, we use the first-order necessary conditions for the optimal $\bv A$:
\begin{equation}
-\left(A^{-\top}\right)_{i,j}+\sum_{n}V_{i,n}^{-1}A_{n,j}+\mathbb{E}_{\bs \epsilon}\left[\frac{\partial\mathrm{L}_n\left(\bs \theta\right)}{\partial\theta_i}\epsilon_j\right]=0\,.
\end{equation}
And in matrix form we obtain:
\begin{align}
-\bv A^{-\top}+\bv V^{-1}\bv A+\mathbb{E}_{\bs \epsilon}\left[\nabla\mathrm{L}_n\left(\bs \theta\right)\bs \epsilon^{\top}\right]=0\,.
\end{align}
And so we obtained \eqref{eq:Gaussian first order necessary conditions}.

\section{Proof of Lemma \ref{lemma}}
\label{app:Proof of Lemma}
The following proof is based on \cite{309879}.
\begin{proof}
Let 
\begin{align}
\bv X = \bv D \bv Q - \frac{1}{2}\bv M \bv T\,, 
\end{align}
such that 
\begin{align}
\bv B &= \bv M + \frac{1}{4}\bv M \bv T \bv T^{\top} \bv M  \\
\bv D &= \bv B^{1/2}\,.
\end{align}
If we compare \eqref{eq:Matrix equation} with its transpose we obtain
\begin{align}
\bv M \bv T\bv X^{\top} = \bv X \bv T^{\top}\bv M    \,,
\end{align}
so we can rewrite \eqref{eq:Matrix equation} as follows,
\begin{align}
\bv X\bv X^{\top} + \frac{1}{2}\bv M \bv T\bv X^{\top} +\frac{1}{2}\bv X \bv T^{\top}\bv M -\bv M = 0\,. \label{eq:equivalent matrix equation}
\end{align}
Next, we can factor \eqref{eq:equivalent matrix equation} as follows,
\begin{align}
\left(\bv X + \frac{1}{2}\bv M \bv T\right)\left(\bv X + \frac{1}{2}\bv T^{\top} \bv M\right)^{\top} = \bv M + \frac{1}{4}\bv M \bv T \bv T^{\top} \bv M\,. 
\end{align}
Since the matrix $\bv B = \bv M + \frac{1}{4}\bv M \bv T \bv T^{\top} \bv M$ is positive definite (PD)
\begin{align}
\bv D^{-1}\left(\bv X + \frac{1}{2}\bv M \bv T\right)\left(\bv X + \frac{1}{2}\bv T^{\top} \bv M\right)\bv D^{-1} = \bv I_N\,,
\end{align}
the equality holds if and only if $\bv Q = \bv D^{-1}\left(\bv X + \frac{1}{2}\bv M \bv T\right)$ is an orthogonal matrix.
In addition, the matrix 
\begin{align}
    \bv X \bv T^{\top}\bv M = \left(\bv D \bv Q - \frac{1}{2}\bv M \bv T\right)\bv T^{\top}\bv M = \bv D \bv Q \bv T^{\top}\bv M - \frac{1}{2}\bv M \bv T \bv T^{\top}\bv M
\end{align}
is symmetric if and only if $\bv D \bv Q \bv T^{\top}\bv M$ is symmetric.
\end{proof}

\section{Proof of Lemma \ref{lemma1}}
\label{app:Proof of Lemma1}
The following proof is based on \cite{309879}.
\begin{proof}
Let 
\begin{align}
\bv  Q = \bv S \bv W^{\top}
\end{align}
such that 
$\bv S, \bv W$ are the left and right singular matrices of the Singular Value Decomposition (SVD) of $\bv D^{-1} \bv M \bv T$.
Then
\begin{align}
    \bv Q^{\top} \bv Q = \bv W \bv S^{\top}\bv S \bv W^{\top} &= \bv I 
\end{align}
and 
\begin{align}
    \bv D \bv Q \bv T^{\top}\bv M &= \bv D \bv S \bv W^{\top} \bv T^{\top}\bv M = \bv D \bv S \bv W^{\top} \bv T^{\top}\bv M \bv D^{-\top}\bv D^{\top} \\
    &\overset{(a)}{=} \bv D \bv S \bv W^{\top} \bv W \bs \Lambda \bv S^{\top}\bv D^{\top} = \bv D \bv S \bs \Lambda \bv S^{\top}\bv D^{\top}
\end{align}
where $(a)$ is because $\bv S, \bv W$ are the left and right singular matrices of the SVD of $\bv D^{-1} \bv M \bv T$ meaning, $\bv D^{-1} \bv M \bv T = \bv S \bs \Lambda \bv W ^{\top}$ and $\bs \Lambda$ is a diagonal matrix. Therefore $\bv Q$ is an orthogonal matrix and $\bv D \bv Q \bv T^{\top}\bv M$ is a symmetric matrix.
\end{proof}

\section{Lemma \ref{lemma2}}
\label{app:lemma2}
In this section, we derive a solution for \eqref{eq:Matrix equation} in the case where $\bv B$ is not invertible (or, more often, has a large condition number).
\begin{lemma}
\label{lemma2}
In this Lemma we use the notations of Lemma \ref{lemma}. Let  $\bv Q = \bv U \bv Z^{\top}$ such that $\bv U, \bv Z$ are the left singular matrices of the Generalized Singular Value Decomposition (GSVD) of $\left(\bv D^{\top}, \bv T ^{\top} \bv M\right)$, respectively. Then $\bv Q$ is an orthogonal matrix and $\bv D \bv Q \bv T^{\top} \bv M$ is a symmetric matrix.
\end{lemma}
The following proof is based on \cite{309879}.
\begin{proof}
Let 
\begin{align}
\bv  Q = \bv S \bv W^{\top}
\end{align}
such that 
$\bv U, \bv Z$ are the left singular matrices of the Generalized Singular Value Decomposition (GSVD) of $\left(\bv D^{\top}, \bv T ^{\top} \bv M\right)$,  respectively.
Then
\begin{align}
    \bv Q^{\top} \bv Q = \bv Z \bv U^{\top}\bv U \bv Z^{\top} &= \bv I 
\end{align}
and 
\begin{align}
    \bv D \bv Q \bv T^{\top}\bv M &= \bv D \bv U \bv Z^{\top} \bv T^{\top}\bv M 
    \overset{(a)}{=} \bv W \bs \Lambda_1 \bv U^{\top} \bv U \bv Z^{\top} \bv Z \bs \Lambda_2 \bv W ^{\top} \\
    &=\bv W \bs \Lambda_1 \bs \Lambda_2 \bv W ^{\top}
\end{align}
where $(a)$ is because  is because $\bv U, \bv Z$ are the left singular matrices of the GSVD of $\left(\bv D^{\top}, \bv T^{\top} \bv M\right)$ meaning, $\bv D^{\top} = \bv U \bs \Lambda_1 \bv W ^{\top}, \bv T^{\top} \bv M = \bv Z \bs \Lambda_2 \bv W ^{\top}$ and $\bs \Lambda_1, \bs \Lambda_2$ are diagonal matrices. Therefore $\bv Q$ is an orthogonal matrix and $\bv D \bv Q \bv T^{\top}\bv M$ is a symmetric matrix.
\end{proof}

\section{Derivation of \eqref{eq:Kronecker-factored necessary conditions}}
\label{app:Derivation of Kronecker-factored necessary conditions}
In this section, we provide additional details on the derivation of \eqref{eq:Kronecker-factored necessary conditions}. The objective function is 
\begin{align}
f\left(\bs \mu,\bs \Sigma\right)&=
\frac{1}{2}\Bigg[\log\frac{\det{\left(\bv V_1\right)}^{p}\det{\left(\bv V_2\right)}^{n}}{\det{\left(\bs \Sigma_1\right)}^{p}\det{\left(\bs \Sigma_2\right)}^{n}}-np 
+\trace\left(\left(\bv V_1\otimes\bv V_2\right)^{-1}(\bs \Sigma_1\otimes\bs \Sigma_2)\right)\\
&\quad +\left(\bv m-\bs \mu\right)^{\top}\left(\bv V_1\otimes\bv V_2\right)^{-1}\left(\bv m-\bs \mu\right)\Bigg] \nonumber \\ & \quad+
\mathbb{E}_{\bs \theta}\left[\mathrm{L}_n\left(\bs \theta\right)\right]\,.
\end{align}
To solve the optimization problem in \eqref{eq:online variational bayes} in the case of Kronecker-factored approximation, we use the deterministic transformation \eqref{eq:Kronecker-factored deterministic transformation}. To calculate the first derivative of the objective function we use the following identities (see Appendix \ref{app:Technical results} for additional details):
\begin{align}
\frac{\partial\mathbb{E}_{\bs \epsilon}\left[\mathrm{L}_n\left(\bs \theta\right)\right]}{\partial A_{i,j}}
&= \mathbb{E}_{\bs \epsilon}\left[\sum_{\ell=1}^{p}\sum_{k=1}^{p}\frac{\partial\mathrm{L}_n\left(\bs \theta\right)}{\partial\theta_{\ell+(i-1)p}}B_{\ell,k}\bs \epsilon_{k+(j-1)p}\right] \\
\frac{\partial\mathbb{E}_{\bs \epsilon}\left[\mathrm{L}_n\left(\bs \theta\right)\right]}{\partial B_{i,j}} &=  \mathbb{E}_{\bs \epsilon}\left[\sum_{\ell=1}^{n}\sum_{k=1}^{n}\frac{\partial\mathrm{L}_n\left(\bs \theta\right)}{\partial\theta_{i+(\ell-1)n}}A_{\ell,k}\bs \epsilon_{j+(k-1)n}\right] \\
\trace\left(\left(\bv V_1\otimes\bv V_2\right)^{-1}(\bs \Sigma_1\otimes\bs \Sigma_2)\right)
&=\trace\left(\bv V_1^{-1}\bs \Sigma_1\right)\trace\left(\bv V_2^{-1}\bs \Sigma_2\right)\,.
\end{align}
We use the first-order necessary conditions for the optimal $\bs \mu$:
\begin{equation}
-\left(\bv V_1\otimes\bv V_2\right)^{-1}\left(\bv m-\bs \mu\right)+\mathbb{E}_{\bs \epsilon}\left[\nabla\mathrm{L}_n\left(\bs \theta\right)\right] = 0\,.
\end{equation}
And so we obtained \eqref{eq:Kronecker-factored necessary conditions}.
We use the first-order necessary conditions for the optimal $\bv A$:
\begin{equation}
-p\left(A^{-\top}\right)_{i,j}+\trace\left(\bv V_2^{-1}\bs \Sigma_2\right)\sum_{k}(V_1)_{i,k}^{-1}A_{k,j}+\mathbb{E}_{\bs \epsilon}\left[\sum_{\ell=1}^{p}\sum_{k=1}^{p}\frac{\partial\mathrm{L}_n\left(\bs \theta\right)}{\partial\theta_{\ell+(i-1)p}}B_{\ell,k} \epsilon_{k+(j-1)p}\right]=0\,.
\end{equation}
And in matrix form we obtain:
\begin{align}
-p\bv A^{-\top}+\trace\left(\bv V_2^{-1}\bs \Sigma_2\right)\bv V_1^{-1}\bv A+\mathbb{E}_{\bs \epsilon}\left[\Psi^{\top}\bv B\Phi\right]=0 \,.
\end{align}
where $\Psi,\Phi\in\mathbb{R}^{p\times n}$ such that $\Vectorization(\Psi)=\nabla\mathrm{L}_n\left(\bs \theta\right)$ and $\Vectorization(\Phi)=\bs \epsilon$. And so we obtained \eqref{eq:Kronecker-factored necessary conditions}.

We use the first-order necessary conditions for the optimal $\bv B$:
\begin{equation}
-n\left(B^{-\top}\right)_{i,j}+\trace\left(\bv V_1^{-1}\bs \Sigma_1\right)\sum_{k}(V_2)_{i,k}^{-1}B_{k,j}+\mathbb{E}_{\bs \epsilon}\left[\sum_{\ell=1}^{n}\sum_{k=1}^{n}\frac{\partial\mathrm{L}_n\left(\bs \theta\right)}{\partial\theta_{i+(\ell-1)n}}A_{\ell,k} \epsilon_{j+(k-1)n}\right]=0\,.
\end{equation}
And in matrix form we obtain:
\begin{align}
-n\bv B^{-\top}+\trace\left(\bv V_1^{-1}\bs \Sigma_1\right)\bv V_2^{-1}\bv B+\mathbb{E}_{\bs \epsilon}\left[\Psi^{\top}\bv A\Phi\right]=0\,.
\end{align}
And so we obtained \eqref{eq:Kronecker-factored necessary conditions}.

\section{Technical results for Appendix \ref{app:Derivation of Kronecker-factored necessary conditions}}
\label{app:Technical results}
In this section, we prove the technical results used in Appendix \ref{app:Derivation of Kronecker-factored necessary conditions}.

Let  $\Delta_{i,j}^{(n)}\in\mathbb{R}^{n\times n}$ such that 
\[
(\Delta_{i,j}^{(n)})_{n,m}=\delta_{n,i}\delta_{m,j}.
\]
We then have
\begin{align}
\frac{\partial\mathbb{E}_{\bs \epsilon}\left[\mathrm{L}_n\left(\bs \theta\right)\right]}{\partial A_{i,j}} &= \mathbb{E}_{\bs \epsilon}\left[\frac{\partial\mathrm{L}_n\left(\bs \theta\right)}{\partial A_{i,j}}\right] \nonumber \\ 
&\eqann{a} \mathbb{E}_{\bs \epsilon}\left[\sum_{\ell=1}^{np}\frac{\partial\mathrm{L}_n\left(\bs \theta\right)}{\partial\theta_{\ell}}\cdot\frac{\partial\theta_{\ell}}{\partial A_{i,j}}\right] \nonumber \\
&\eqann{b} \mathbb{E}_{\bs \epsilon}\left[\sum_{\ell=1}^{np}\frac{\partial\mathrm{L}_n\left(\bs \theta\right)}{\partial\theta_{\ell}}\cdot\frac{\partial (\mu_{\ell} + \sum_k (\bv A\otimes \bv B)_{\ell,k}\epsilon_k)}{\partial A_{i,j}}\right] \nonumber \\
&\eqann{c} \mathbb{E}_{\bs \epsilon}\left[\sum_{\ell=1}^{np}\frac{\partial\mathrm{L}_n\left(\bs \theta\right)}{\partial\theta_{\ell}}\sum_{k=1}^{np}\left(\Delta_{i,j}^{(n)} \otimes \bv B\right)_{\ell,k}\bs \epsilon_k\right]\nonumber \\
&\eqann{d} \mathbb{E}_{\bs \epsilon}\left[\sum_{\ell=(i-1)p+1}^{ip}\frac{\partial\mathrm{L}_n\left(\bs \theta\right)}{\partial\theta_{\ell}}\sum_{k=(j-1)p+1}^{jp}\left(\Delta_{i,j}^{(n)} \otimes \bv B\right)_{\ell,k}\bs \epsilon_k\right]\nonumber \\
&= \mathbb{E}_{\bs \epsilon}\left[\sum_{\ell=1}^{p}\sum_{k=1}^{p}\frac{\partial\mathrm{L}_n\left(\bs \theta\right)}{\partial\theta_{\ell+(i-1)p}}B_{\ell,k}\bs \epsilon_{k+(j-1)p}\right],
\end{align}
where
\begin{itemize}
\item \eqannref{a} is because $\bs \theta$ is a vector of length $n\cdot p$ and by the chain rule for derivatives; 
\item \eqannref{b} holds since $\theta_{\ell} = \mu_{\ell} + \sum_{k=1}^{np} \left(\bv A\otimes \bv B\right)_{\ell,k} \epsilon_k$;
\item \eqannref{c} is by definition of $\Delta_{i,j}^{(n)}$, and since we differentiate by $A_{i,j}$;
\item \eqannref{d} is since $(\Delta_{i,j}^{(n)} \otimes \bv B)_{\ell,k} \neq 0$ if $(i-1)p+1 \leq \ell \leq ip$ or if $(j-1)p+1 \leq k \leq jp$.
\end{itemize}

Let  $\Delta_{i,j}^{(p)}\in\mathbb{R}^{p\times p}$ such that 
\[
(\Delta_{i,j}^{(p)})_{n,m}=\delta_{n,i}\delta_{m,j}.
\]
We then have
\begin{align}
\frac{\partial\mathbb{E}_{\bs \epsilon}\left[\mathrm{L}_n\left(\bs \theta\right)\right]}{\partial B_{i,j}} &= \mathbb{E}_{\bs \epsilon}\left[\frac{\partial\mathrm{L}_n\left(\bs \theta\right)}{\partial B_{i,j}}\right] \nonumber \\ 
&\eqann{a} \mathbb{E}_{\bs \epsilon}\left[\sum_{\ell=1}^{np}\frac{\partial\mathrm{L}_n\left(\bs \theta\right)}{\partial\theta_{\ell}}\cdot\frac{\partial\theta_{\ell}}{\partial B_{i,j}}\right] \nonumber \\
&\eqann{b} \mathbb{E}_{\bs \epsilon}\left[\sum_{\ell=1}^{np}\frac{\partial\mathrm{L}_n\left(\bs \theta\right)}{\partial\theta_{\ell}}\cdot\frac{\partial (\mu_{\ell} + \sum_k (\bv A\otimes \bv B)_{\ell,k}\epsilon_k)}{\partial B_{i,j}}\right] \nonumber \\
&\eqann{c} \mathbb{E}_{\bs \epsilon}\left[\sum_{\ell=1}^{np}\frac{\partial\mathrm{L}_n\left(\bs \theta\right)}{\partial\theta_{\ell}}\sum_{k=1}^{np}\left(\bv A \otimes \Delta_{i,j}^{(p)}\right)_{\ell,k}\bs \epsilon_k\right]\nonumber \\
&\eqann{d} \mathbb{E}_{\bs \epsilon}\left[\sum_{\ell=1}^{n}\sum_{k=1}^{n}\frac{\partial\mathrm{L}_n\left(\bs \theta\right)}{\partial\theta_{i+(\ell-1)n}}A_{\ell,k}\bs \epsilon_{j+(k-1)n}\right],
\end{align}
where
\begin{itemize}
\item \eqannref{a} is because $\bs \theta$ is a vector of length $n\cdot p$ and by the chain rule for derivatives; 
\item \eqannref{b} holds since $\theta_{\ell} = \mu_{\ell} + \sum_{k=1}^{np} \left(\bv A\otimes \bv B\right)_{\ell,k} \epsilon_k$;
\item \eqannref{c} is by definition of $\Delta_{i,j}^{(p)}$, and since we differentiate by $B_{i,j}$;
\item \eqannref{d} is since $(\bv A \otimes \Delta_{i,j}^{(p)})_{\ell,k} \neq 0$ if $k\mod n = j$ and if $\ell\mod n = i$.
\end{itemize}

\section{Derivation of the fixed-point equations for the matrix variate Gaussian}
\label{app:fixed-point equations for the  matrix variate Gaussian}
In this section, we provide additional details on the derivation of the fixed-point equations for the the matrix variate Gaussian.
\begin{align}
&\bs \mu = \bv m-\left(\bv V_1\otimes \bv V_2\right)\mathbb{E}_{\bs \epsilon}\left[\nabla\mathrm{L}_n\left(\bs \theta\right)\right] \\
&\bv A\bv A^{\top}+ \left(\frac{p}{\trace\left(\bv V_2^{-1}\bs \Sigma_2\right)}\right)\bv V_1\frac{1}{p}\mathbb{E}_{\bs \epsilon}\left[\Psi^{\top}\bv B\Phi\right]\bv A^{\top}-\left(\frac{p}{\trace\left(\bv V_2^{-1}\bs \Sigma_2\right)}\right)\bv V_1=0 \\
&\bv B\bv B^{\top}+\left(\frac{n}{\trace\left(\bv V_1^{-1}\bs \Sigma_1\right)}\right)\bv V_2\frac{1}{n}\mathbb{E}_{\bs \epsilon}\left[\Psi\bv A\Phi^{\top}\right]\bv B^{\top}-\left(\frac{n}{\trace\left(\bv V_1^{-1}\bs \Sigma_1\right)}\right)\bv V_2=0 
\end{align}
Lemma \ref{lemma1} reveals the fixed-point equations
\begin{align}
&\bs \mu = \bv m-\left(\bv V_1\otimes \bv V_2\right)\mathbb{E}_{\bs \epsilon}\left[\nabla\mathrm{L}_n\left(\bs \theta\right)\right] \\
&\bv A = \bv D_1 \bv Q_1 - \frac{1}{2}\left(\frac{p}{\trace\left(\bv V_2^{-1}\bs \Sigma_2\right)}\right)\bv V_1 \frac{1}{p}\mathbb{E}_{\bs \epsilon}\left[\Psi^{\top}\bv B\Phi\right]\\
&\bv B = \bv D_2 \bv Q_2 - \frac{1}{2}\left(\frac{n}{\trace\left(\bv V_1^{-1}\bs \Sigma_1\right)}\right)\bv V_2 \frac{1}{n}\mathbb{E}_{\bs \epsilon}\left[\Psi\bv A\Phi^{\top}\right]
\end{align}

\section{Proof of Theorem \ref{theorem}}
\label{appendix:Proof of Theorem}
\begin{proof}
We define $\theta_{j}=\mu_{j}+\epsilon_{j}\sigma_{j}$ where  $\epsilon_{j}\sim\mathcal{N}\left(0,1\right)$.
According to the smoothing theorem, the following holds
\begin{equation}
\mathbf{\mathbb{E}}_{\epsilon}\left[\frac{\partial \mathrm{L}_n\left(\bs \theta\right)}{\partial\theta_{i}}\epsilon_{i}\right]=\mathbb{E}_{\epsilon_{j\neq i}}\left[\mathbf{\mathbb{E}}_{\epsilon_{i}}\left[\left.\frac{\partial \mathrm{L}_n\left(\bs \theta\right)}{\partial\theta_{i}}\epsilon_{i}\right|\epsilon_{j\neq i}\right]\right]\,.
\end{equation}
The conditional expectation is:
\begin{equation}
\mathbf{\mathbb{E}}_{\epsilon_{i}}\left[\left.\frac{\partial \mathrm{L}_n\left(\bs \theta\right)}{\partial\theta_{i}}\epsilon_{i}\right|\epsilon_{j\neq i}\right]=\intop_{-\infty}^{\infty}\frac{\partial \mathrm{L}_n\left(\bs \theta\right)}{\partial\theta_{i}}\epsilon_{i}f_{\epsilon_{i}}\left(\epsilon_{i}\right)d\epsilon_{i}\,,
\end{equation}
where $f_{\epsilon_{i}}$ is the probability density function of a standard normal distribution. Since $f_{\epsilon_{i}}$ is an even function
\begin{align}
& \mathbf{\mathrm{E}}_{\epsilon_{i}}\left[\left.\frac{\partial \mathrm{L}_n\left(\bs \theta\right)}{\partial\theta_{i}}\epsilon_{i}\right|\epsilon_{j\neq i}\right] 
=  \intop_{0}^{\infty}\frac{\partial \mathrm{L}_n\left(\mu_{i}+\epsilon_{i}\sigma_{i},\theta_{-i}\right)}{\partial\theta_{i}}\epsilon_{i}f_{\epsilon_{i}}\left(\epsilon_{i}\right)d\epsilon_{i} -\intop_{0}^{\infty}\frac{\partial \mathrm{L}_n\left(\mu_{i}-\epsilon_{i}\sigma_{i},\theta_{-i}\right)}{\partial\theta_{i}}\epsilon_{i}f_{\epsilon_{i}}\left(\epsilon_{i}\right)d\epsilon_{i}\,. \label{eq:expectation}
\end{align}

Now, since $\mathrm{L}_n\left(\bs \theta\right)$ is strongly convex function
with parameter $m_n>0$ and continuously differentiable function over
$\mathbb{R}^{d}$, the following holds $\forall\bs \theta_{1},\bs \theta_{2}\in\mathbb{R}^{d}$:
\begin{equation}
\left(\nabla \mathrm{L}_n\left(\bs \theta_{1}\right)-\nabla \mathrm{L}_n\left(\bs \theta_{2}\right)\right)^{T}\left(\bs \theta_{1}-\bs \theta_{2}\right)\geq m_n\left\| \bs \theta_{1}-\bs \theta_{2}\right\|_2^2\,.
\end{equation}
For $\bs \theta_{1},\bs \theta_{2}$ such that
\begin{align}
\left(\bs \theta_{1}\right)_{j}=\begin{cases}
\left(\bs \theta_{2}\right)_{j}, & j\neq i\\
\mu_{i}+\epsilon_{i}\sigma_{i}, & j=i\,,
\end{cases}
\end{align}
\begin{align}
\left(\bs \theta_{2}\right)_{j}=\begin{cases}
\left(\bs \theta_{1}\right)_{j}, & j\neq i\\
\mu_{i}-\epsilon_{i}\sigma_{i}, & j=i\,,
\end{cases}
\end{align}
the following holds:
\begin{equation}
\left(\frac{\partial \mathrm{L}\left(\bs \theta_{1}\right)}{\partial\theta_{i}}-\frac{\partial \mathrm{L}\left(\bs \theta_{2}\right)}{\partial\theta_{i}}\right)\epsilon_{i}\geq 2m_n\sigma_i\epsilon_{i}^2\,.
\end{equation}
Therefore, substituting this inequality into \eqref{eq:expectation}, we obtain:
\begin{equation}
\mathbf{\mathbb{E}}_{\epsilon}\left[\frac{\partial \mathrm{L}\left(\bs \theta\right)}{\partial\theta_{i}}\epsilon_{i}\right]\geq m_n\sigma_i>0\,.
\end{equation}
\end{proof}

\section{FOO-VB Algorithm: Some Special Cases}
\label{app:algorithms}
In this section, we present the FOO-VB algorithms for the matrix variate Gaussian and diagonal Gaussian variants.
\subsection{Matrix variate Gaussian}
In the case of matrix variate Gaussian, one can write the deterministic transformation  
\begin{align}
\bs \theta =\bs \mu+\left(\bv A\otimes\bv B\right)\bs \epsilon
\end{align}
in matrix form
\begin{align}
\bv W =\bv M+\bv B \bs \Phi \bv A ^{\top} \,,
\end{align}
where $\bv W, \bs \Phi \in\mathbb{R}^{d_2\times d_1}$ and $\Vectorization{\left(\bv W\right)} = \bs \theta, \Vectorization{\left(\bs \Phi\right)} = \bs \epsilon$. 
The deterministic transformation and the posterior update rule for the matrix variate Gaussian in the matrix form can be found in Algorithm \ref{algo:KFAC Gaussian}. 
\subsection{Diagonal Gaussian}
The deterministic transformation and the posterior update rule for the diagonal Gaussian can be found in Algorithm \ref{algo:Diagonal Gaussian}. 

\begin{algorithm}
\caption{Methods for the matrix variate Gaussian version}
\begin{description}
\item [{\textsc{Transform}$\left(\bs \epsilon,\phi = \left(\bv M, \bv A, \bv B\right)\right):$}] ~
\begin{description}
\item $\bv W =\bv M+\bv B\bs \Phi \bv A^{\top}$
\item $\bs \Sigma_1 = \bv A\bv A^\top $
\item $\bs \Sigma_2 = \bv B\bv B^\top $
\end{description}
\item [{\textsc{Update}$\left(\phi_{n-1} = \left(\bv M, \bv A, \bv B\right), \bs \epsilon^{(1..K)}, \bv g^{(1..K)}\right)$:}] ~
\begin{description}
\item $\Vectorization(\Psi^{(k)})=\bv g^{(k)}$
\item $\Vectorization(\Phi^{(k)})=\bs \epsilon^{(k)}$
\item $\bar{\bv E}_{1} = \frac{1}{K}\sum_{k=1}^{K}\bs \Psi^{(k)}$
\item $\bar{\bv E}_{2} = \frac{1}{K}\sum_{k=1}^{K}\frac{1}{p}\bs \Psi^{(k)\top}\bv B \bs \Phi^{(k)}$
\item $\bar{\bv E}_{3} = \frac{1}{K}\sum_{k=1}^{K}\frac{1}{n}\bs \Psi^{(k)\top}\bv A \bs \Phi^{(k)}$
\item $\bv M \xleftarrow{} \bv M-\bv B \bv B^{\top} \bar{\bv E}_{1} \bv A \bv A^{\top}$
\item $\bv A \xleftarrow{} \bv X \text{ such that } \bv X\bv X^{\top}+ \bv A \bv A^{\top} \bar{\bv E}_{2} \bv X^{\top}-\bv A \bv A^{\top}=0 $
\item $\bv B \xleftarrow{} \bv X \text{ such that } \bv X\bv X^{\top}+ \bv B \bv B^{\top} \bar{\bv E}_{3} \bv X^{\top}-\bv B \bv B^{\top}=0 $
\item (The matrix equations are solved using Lemma \ref{lemma1})
\label{algo:KFAC Gaussian}
\end{description}
\end{description}
\end{algorithm}

\begin{algorithm}
\caption{Methods for the diagonal Gaussian version}
\begin{description}
\item [{\textsc{Transform}$\left(\bs \epsilon,\phi = \left(\bs \mu, \bs \sigma\right)\right):$}] ~
\begin{description}
\item $\theta_i =\mu_i+\sigma_i\epsilon_i$
\end{description}
\item [{\textsc{Update}$\left(\phi_{n-1} = \left(\bs \mu, \bs \sigma\right), \bs \epsilon^{(1..K)}, \bv g^{(1..K)}\right)$:}] ~
\begin{description}
\item $\left(\bar{\bv E}_{1}\right)_i = \frac{1}{K}\sum_{k=1}^{K} \bv g^{(k)}_{i}$
\item $\left(\bar{\bv E}_{2}\right)_i = \frac{1}{K}\sum_{k=1}^{K} \bv g^{(k)}_{i} \bs \epsilon^{(k)}_{i}$
\item $\mu_{i} \xleftarrow{} \mu_{i} -\sigma_i^2 (\bar{\bv E}_{1})_{i}$
\item $\sigma_{i} \xleftarrow{} \sigma_{i}\sqrt{1+\left(\frac{1}{2}\sigma_{i}(\bar{\bv E}_{2})_{i}\right)^{2}} -\frac{1}{2}\sigma_{i}^{2}(\bar{\bv E}_{2})_{i}$
\end{description}
\label{algo:Diagonal Gaussian}
\end{description}
\end{algorithm}

\section{Implementation details}
\label{appendix:Implementation details}
For the matrix variate version of FOO-VB, we initialize the weights by sampling from a Gaussian distribution with zero mean and a variance of $ 2/(n_{\mathrm{input}}+2)$. We do so by sampling the mean of the wights from a Gaussian distribution, such that
\begin{align}
\bv M_{i,j} \sim \mathcal{N}\left(0,\frac{2\alpha}{n_\mathrm{input}+2}\right)\,,
\end{align}
where $\alpha \in (0,1)$.
In addition, we sample the diagonal elements of the matrices $\bv A, \bv B$ form a Gaussian distribution, such that
\begin{align}
\epsilon_i\bv A_{i,i} &\sim \mathcal{N}\left(0,\sqrt{\frac{2(1-\alpha)}{n_\mathrm{input}+2}}\right)\\
\epsilon_i\bv B_{i,i} &\sim \mathcal{N}\left(0,\sqrt{\frac{2(1-\alpha)}{n_\mathrm{input}+2}}\right)\,,
\end{align}
where the non-diagonal elements are initialized to zero.
We use $2500$ Monte Carlo samples to estimate the expected gradient during training, and average the accuracy of $2500$ sampled networks during testing, unless stated otherwise.

For the diagonal version of FOO-VB, we initialize the mean of the weights $\mu$ by sampling from a Gaussian
distribution with a zero mean and a variance of $2/(n_{\mathrm{input}}+n_{\mathrm{output}})$, unless stated otherwise.
We use $10$ Monte Carlo samples to estimate the expected gradient during training, and set the weights to the learned mean testing.

\subsection{Discrete permuted MNIST}
We use a fully connected neural network
with 2 hidden layers of 100 width, ReLUs as activation functions
and softmax output layer with 10 units. We used mini-batch of size 128 and 20 epochs
We conducted a hyper-parameter search for all algorithms and present the best test accuracy for 10 tasks, with $LR \in \{0.1, 0.01, 0.001, 0.0001\}$ and regularization coefficient $\in \{250, 150, 10, 0.1, 0.02\}$ (see tables \ref{table:hyper_param_search - EWConline}, \ref{table:hyper_param_search - MAS} and \ref{table:hyper_param_search - Adam&SGD&Adagrad}). For FOO-VB we used $\alpha = 0.5$ in the matrix variate version, and 
$\sigma_{init} = 0.047$ in the diagonal version.
For Online EWC we used LR of 0.001 and regularization coefficient of 10.0 and for MAS we used LR of 0.001 and regularization coefficient of 0.1.

\subsection{Continuous permuted MNIST}
\begin{figure}[ht]
\begin{center}
\centerline{\includegraphics[width= 0.7\columnwidth]{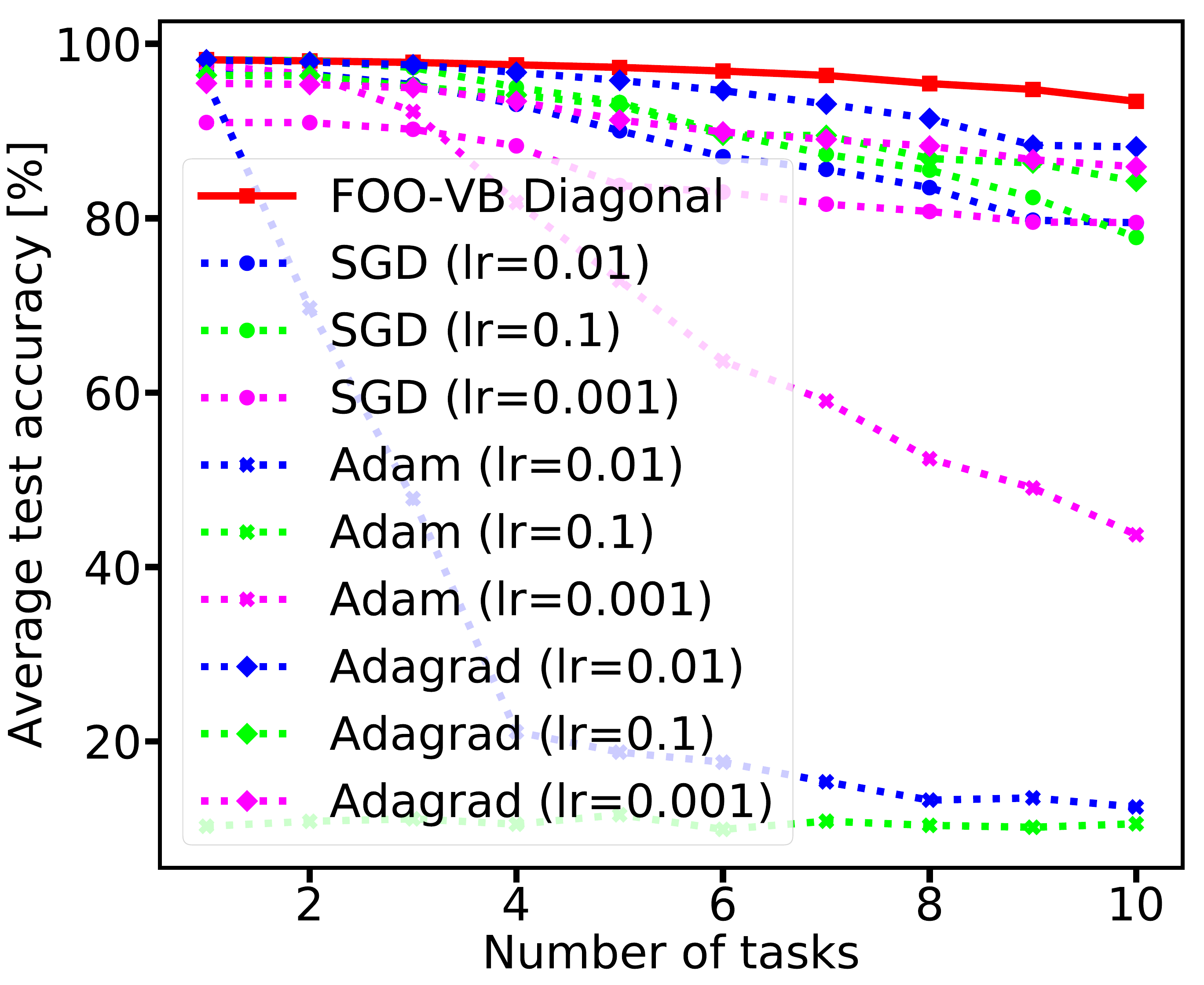}}
\caption{Hyper-parameters search for continuous permuted MNIST experiment.}
\label{cont_permuted_hyperparameter_search}
\vspace{-1em}
\end{center}
\end{figure}

We use a fully connected network with 2 hidden layers of width 200.
Following \cite{hsu2018re}, the original MNIST images are padded with zeros to match size of $32 \times 32$. The batch size is 128, and we sample with replacement due to the properties of the continuous scenario (no definition for epoch as task boundaries are undefined).
We conducted a hyper-parameter search for all algorithms and present the best test accuracy for 10 tasks, see Figure \ref{cont_permuted_hyperparameter_search}.
For FOO-VB we used $\alpha = 0.6$ in the matrix variate version, and 
$\sigma_{init} = 0.06$ in the diagonal version.
For Online EWC and MAS we used the following combinations of hyper-parameters: LR of 0.01 and 0.0001, regularization coefficient of 10, 0.1, 0.001, 0.0001, and optimizer SGD and Adam. The best results for online EWC were achieved using LR of 0.01, regularization coefficient of 0.1 and SGD optimizer.
For MAS, best results achieved using LR of 0.01, regularization coefficient of 0.001 and SGD optimizer.

\subsection{Vision datasets mix}
We followed the experiment as described in \cite{ritter2018online}. We use a batch size of 64, and we normalize the datasets to have zero mean and unit variance. The network architecture is LeNet like with 2 convolution layers with 20 and 50 channels and kernel size of 5, each convolution layer is followed by a Relu activations function and max pool, and the two layers are followed with fully connected layer of size 500 before the last layer.
SGD baseline was trained with a constant learning rate of 0.001 and ADAM used $\epsilon=10^{-8}$, LR of 0.001 and $(\beta_1,\beta_2)=(0.9, 0.999)$. FOO-VD trained with initial STD of 0.02 and batch size 64.
\section{Complexity}
\label{appendix:complexity discussion}

The diagonal version of FOO-VB requires $\times 2$ more parameters compared to SGD, as it stores both the mean and the STD per weight.
In terms of time complexity, the major difference between SGD and FOO-VB arises from the estimation of the expected gradients using Monte Carlo samples during training.
Since those Monte Carlo samples are completely independent the algorithm is embarrassingly parallel.

Specifically, given a mini-batch: for each Monte Carlo sample, FOO-VB generates a random network using $\mu$ and $\sigma$, then making a forward-backward pass with the randomized weights.

Two main implementation methods are available (using 10 Monte Carlo samples as an example):
\begin{enumerate}
    \item
    Producing the (10) Monte Carlo samples sequentially, thus saving only a single randomized network in memory at a time (decreasing memory usage, increasing runtime).
    \item
    Producing the (10) Monte Carlo samples in parallel, thus saving (10) randomized networks in memory (increasing memory usage, decreasing runtime).
\end{enumerate}

We analyzed how the number of Monte Carlo iterations affects the runtime on Continuous permuted MNIST using the first method of implementation (sequential MC samples). The results, reported in Table \ref{timinganalysis}, show that runtime is indeed a linear function of the number of MC iterations.

\begin{table}
\caption{Average runtime of a single training epoch with different numbers of Monte Carlo samples. The MC iterations have linear effect on runtime for classification, and almost no effect in the continuous task-agnostic experiment, probably due to implementation specifics. Using less MC iterations does not affect accuracy significantly. Accuracy reported in the table is from the continuous experiment (Fig \ref{fig:cont_permuted_mnist_results}).}
\begin{center}
\begin{tabular}{c|c|c|c|c}
\textbf{Experiment} & \textbf{MC iterations} & \textbf{Accuracy} & \textbf{Iteration}& \textbf{Vs. SGD} \\
& &  & \textbf{runtime} [seconds] &
\tabularnewline
\hline
Continuous & SGD & 77.79 \% & 0.0024 & $\times$1 \tabularnewline
\cline{2-5}
task-agnostic & 2 (FOO-VB) & 92 \% & 0.0075 & $\times$ 3.12 \tabularnewline
\cline{2-5}
(Fig \ref{fig:cont_permuted_mnist_results}) & 10 (FOO-VB) & 93.40\% & 0.0287 & $\times$ 11.95
\label{timinganalysis}
\end{tabular}
\end{center}
\end{table}

For the matrix variate version of FOO-VB, the main bottleneck is the SVD operation, with the following breakdown:
\begin{itemize}
    \item A single MC iteration takes 0.002 seconds
    \item A single iteration (including all MC iterations and the matrix updates) takes 0.68 seconds (with 10 MC iterations)
    \item Each SVD takes 0.22 seconds, in this case we have two SVDs, which takes 0.44 seconds - about $\frac{2}{3}$ of the iteration runtime.
\end{itemize}{}

In the experiments we used a single GPU (GeForce GTX 1080 Ti).

\begin{table}
\caption{Hyper-parameter search results on discrete permuted MNIST - EWCOnline}
\begin{center}
\begin{tabular}{c|c|c|c}
\textbf{Algorithm} & \textbf{LR} & \textbf{Regularization coef.} & \textbf{Accuracy} 
\tabularnewline
\hline
EWConline & 0.01 & 250.0 & 38.47 \\ 
EWConline & 0.01 & 150.0 & 48.33 \\ 
EWConline & 0.01 & 10.0 & 81.33 \\ 
EWConline & 0.01 & 0.1 & 42.46 \\ 
EWConline & 0.01 & 0.02 & 39.40 \\ 
EWConline & 0.0001 & 250.0 & 46.70 \\ 
EWConline & 0.0001 & 150.0 & 55.58 \\ 
EWConline & 0.0001 & 10.0 & 84.03 \\ 
EWConline & 0.0001 & 0.1 & 58.68 \\ 
EWConline & 0.0001 & 0.02 & 55.17 \\ 
EWConline & 0.001 & 250.0 & 46.20 \\ 
EWConline & 0.001 & 150.0 & 54.19 \\ 
EWConline & 0.001 & 10.0 & 85.06 \\ 
EWConline & 0.001 & 0.1 & 52.13 \\ 
EWConline & 0.001 & 0.02 & 40.48 \\ 
EWConline & 0.1 & 250.0 & 19.32 \\ 
EWConline & 0.1 & 150.0 & 29.05 \\ 
EWConline & 0.1 & 10.0 & 10.04 \\ 
EWConline & 0.1 & 0.1 & 10.49 \\ 
EWConline & 0.1 & 0.02 & 9.688 
\label{table:hyper_param_search - EWConline}
\end{tabular}
\end{center}
\end{table}

\begin{table}
\caption{Hyper-parameter search results on discrete permuted MNIST - MAS}
\begin{center}
\begin{tabular}{c|c|c|c}
\textbf{Algorithm} & \textbf{LR} & \textbf{Regularization coef.} & \textbf{Accuracy} 
\tabularnewline
\hline
MAS & 0.01 & 250.0 & 29.73 \\ 
MAS & 0.01 & 150.0 & 29.85 \\ 
MAS & 0.01 & 10.0 & 44.20 \\ 
MAS & 0.01 & 0.1 & 63.50 \\ 
MAS & 0.01 & 0.02 & 63.46 \\ 
MAS & 0.0001 & 250.0 & 49.57 \\ 
MAS & 0.0001 & 150.0 & 54.13 \\ 
MAS & 0.0001 & 10.0 & 78.03 \\ 
MAS & 0.0001 & 0.1 & 83.49 \\ 
MAS & 0.0001 & 0.02 & 75.32 \\
MAS & 0.001 & 250.0 & 46.41 \\ 
MAS & 0.001 & 150.0 & 50.64 \\ 
MAS & 0.001 & 10.0 & 72.59 \\ 
MAS & 0.001 & 0.1 & 84.56 \\ 
MAS & 0.001 & 0.02 & 82.63 \\ 
MAS & 0.1 & 250.0 & 8.948 \\ 
MAS & 0.1 & 150.0 & 9.772 \\ 
MAS & 0.1 & 10.0 & 13.64 \\ 
MAS & 0.1 & 0.1 & 21.45 \\ 
MAS & 0.1 & 0.02 & 19.49 
\label{table:hyper_param_search - MAS}
\end{tabular}
\end{center}
\end{table}

\begin{table}
\caption{Hyper-parameter search results on discrete permuted MNIST - Adam, SGD and Adagrad}
\begin{center}
\begin{tabular}{c|c|c|c}
\textbf{Algorithm} & \textbf{LR} & \textbf{Regularization coef.} & \textbf{Accuracy} 
\tabularnewline
\hline
Adam & 0.0001 & 0.0 & 52.64 \\ 
Adam & 0.01 & 0.0 & 10.69 \\ 
Adam & 0.1 & 0.0 & 10.31 \\ 
Adam & 0.001 & 0.0 & 27.20 \\ 
SGD & 0.01 & 0.0 & 66.18 \\ 
SGD & 0.0001 & 0.0 & 71.17 \\ 
SGD & 0.001 & 0.0 & 76.94 \\ 
SGD & 0.1 & 0.0 & 36.97 \\ 
Adagrad & 0.1 & 0.0 & 51.48 \\ 
Adagrad & 0.001 & 0.0 & 82.42 \\ 
Adagrad & 0.0001 & 0.0 & 74.14 \\ 
Adagrad & 0.01 & 0.0 & 75.98
\label{table:hyper_param_search - Adam&SGD&Adagrad}
\end{tabular}
\end{center}
\end{table}

\section{5000 epochs training}
\label{appendix:5000 epochs training}
We turn to a MNIST classification experiment to demonstrate the convergence of the log-likelihood cost function and the histogram of STD values. We train a fully connected neural network with two hidden layers and layer width of 400 for 5000 epochs.

\begin{figure}[ht]
\begin{center}
\centerline{\includegraphics[width= 0.7\columnwidth]{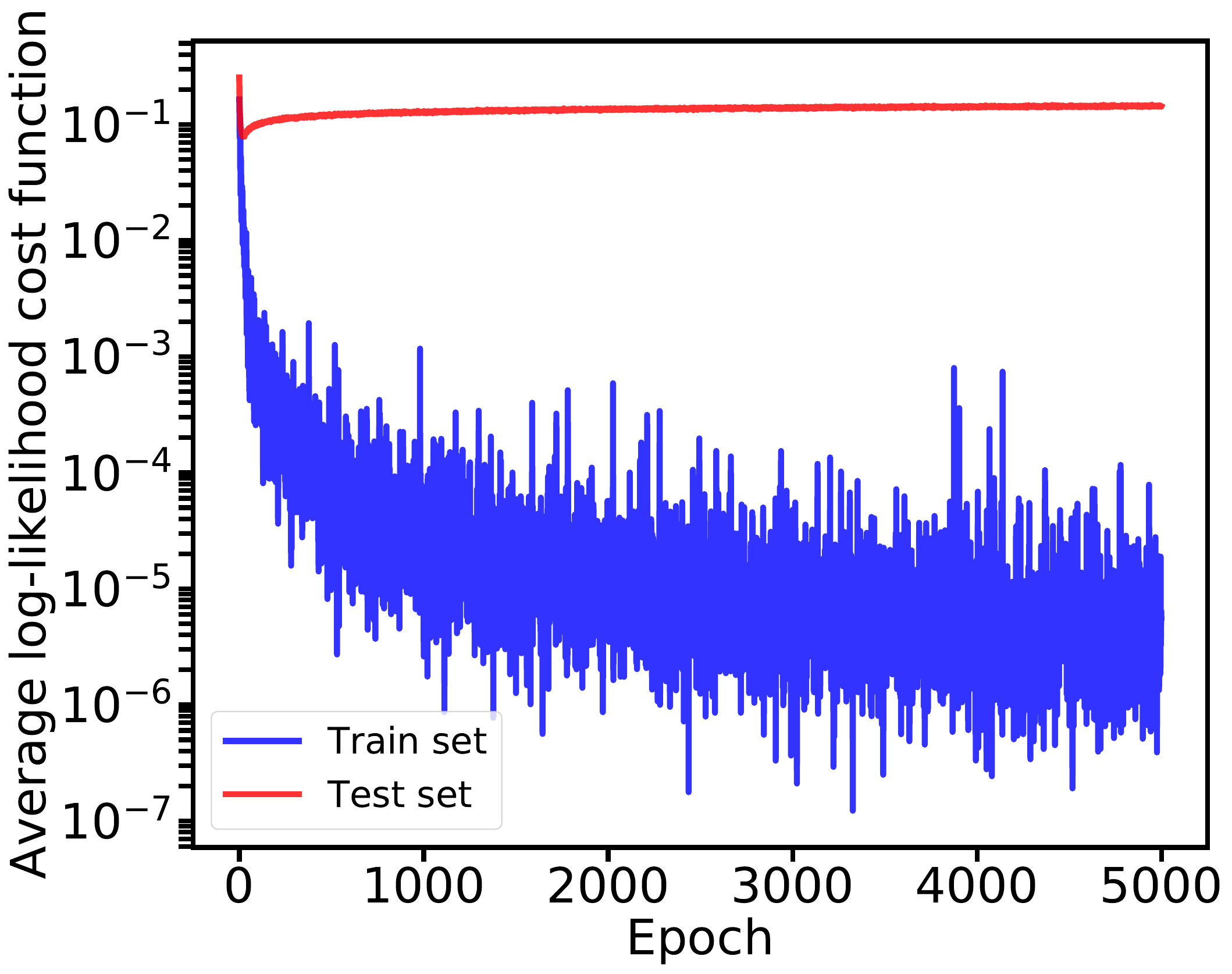}}
\caption{Average log-likelihood cost function of the train set and the test set - layer width 400.}
\label{loss_mnist}
\vspace{-1em}
\end{center}
\end{figure}

Figure \ref{loss_mnist} shows the log-likelihood cost function of the training set and the test set. As can be seen, the log-likelihood cost function on the training set decreases during the training process and converges to a low value. Thus, FOO-VB does not experience underfitting and over-pruning as was shown by~\cite{trippe2018overpruning} for BBB \citep{blundell2015weight}.

\begin{figure}[ht]
\begin{center}
\centerline{\includegraphics[width= 0.7\columnwidth]{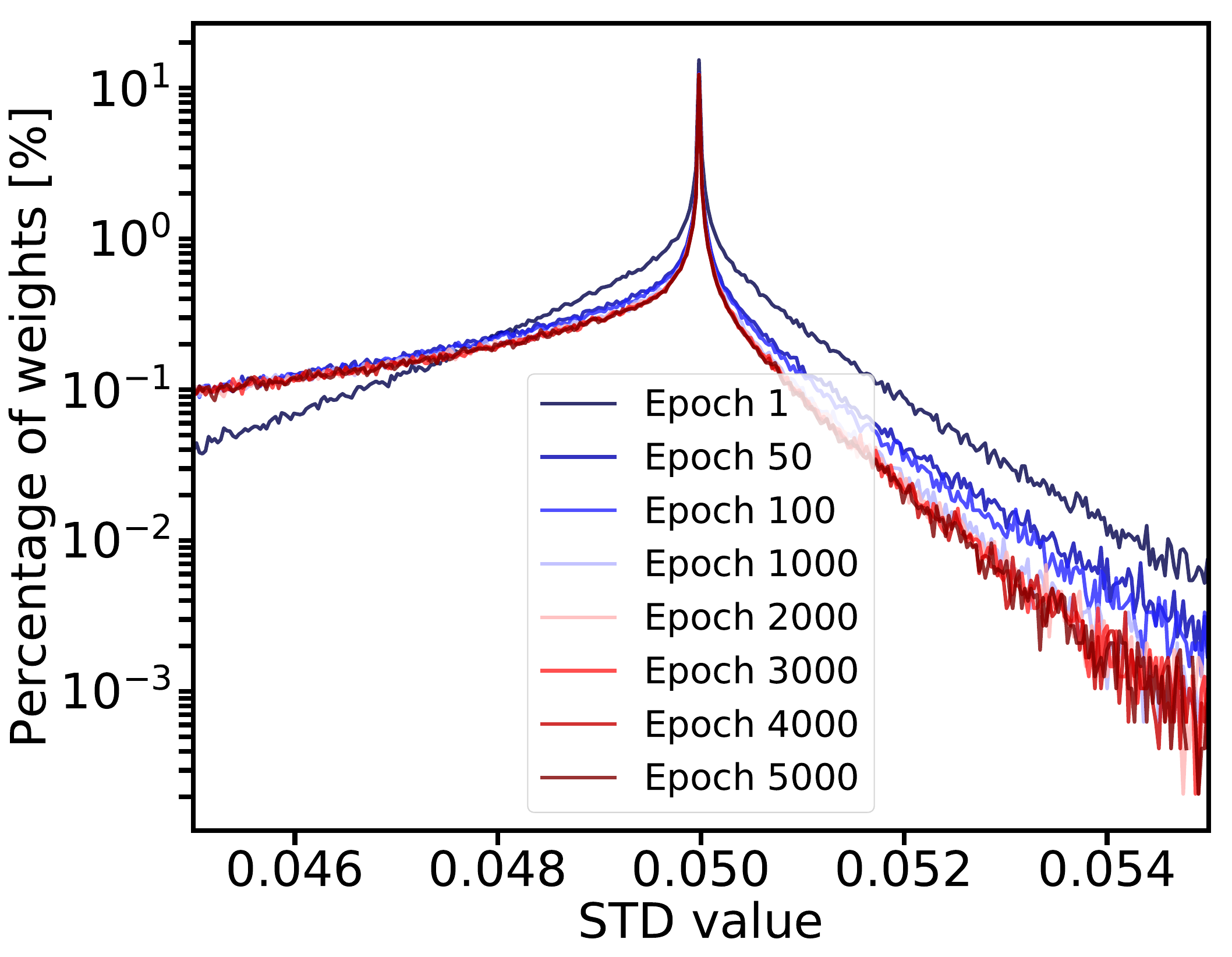}}
\caption{Histogram of STD values, the initial STD value is 0.05.} 
\label{convergence_test}
\end{center}
\end{figure}

Figure \ref{convergence_test} shows the histogram of STD values during the training process.
As can be seen, the histogram of STD values converges. This demonstrates that $\sigma_{i}$ does not collapse to zero even after 5000 epochs. 

\begin{figure}[ht]
\begin{center}
\centerline{\includegraphics[width= 0.7\columnwidth]{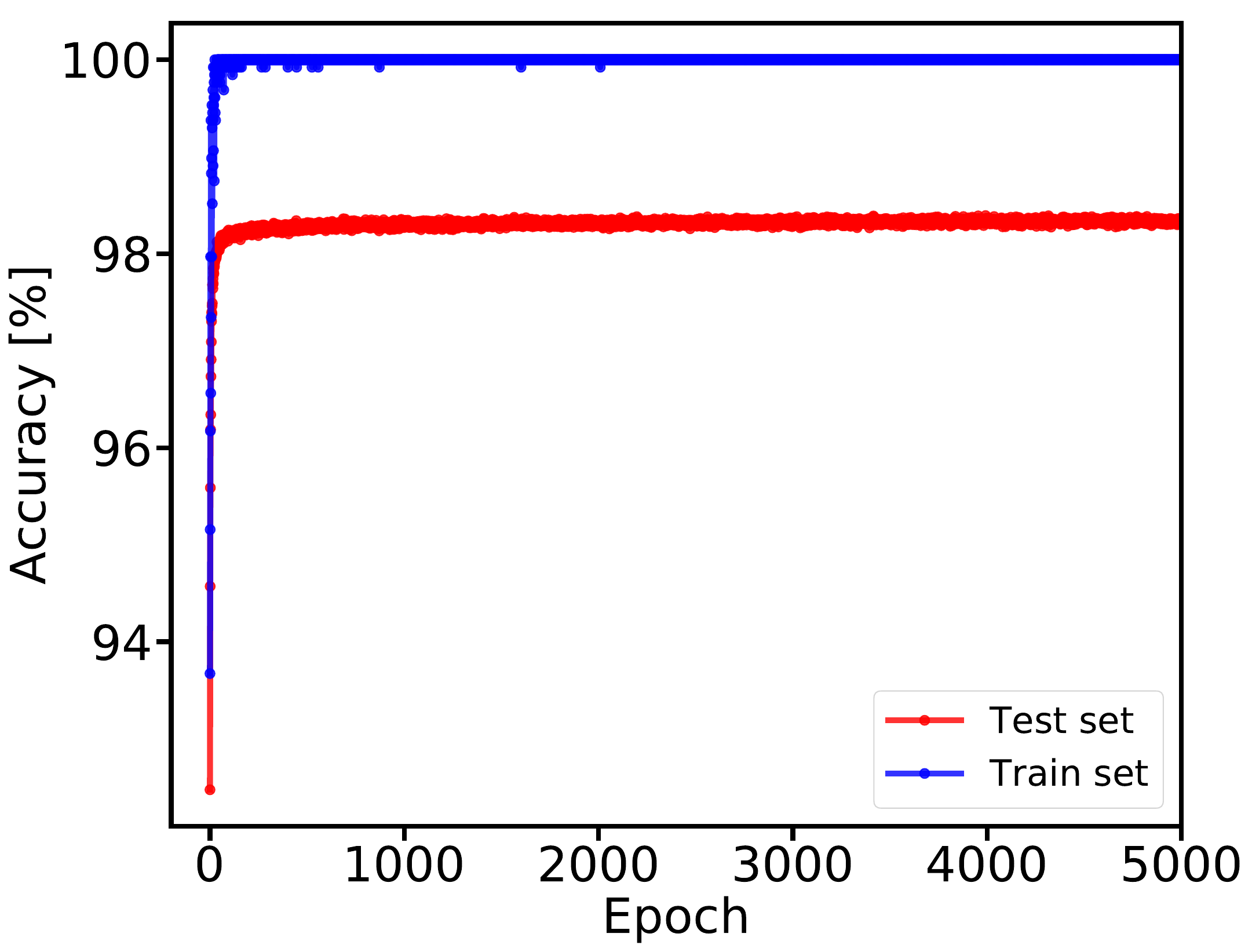}}
\caption{Test accuracy and train accuracy - layer width 400.}
\label{accuracy_mnist}
\vspace{-1em}
\end{center}
\end{figure}

Figure \ref{accuracy_mnist} shows the learning curve of the train set and the test set. As can be seen, the test accuracy does not drop even if we continue to train for 5000 epochs.

\section{Task-aware continual learning on vision datasets}
\label{appendix:vision dataset}
We followed \cite{ritter2018online} and challenged our algorithm with the vision datasets experiment. In this experiment, we train sequentially on MNIST, notMNIST, \footnote{\normalsize Originally published at \\ http://yaroslavvb.blogspot.co.uk/2011/09/notmnist-dataset.html and downloaded from https://github.com/davidflanagan/notMNIST-to-MNIST} FashionMNIST, SVHN and CIFAR10 \citep{lecun1998gradient,xiao2017fashion,netzer2011reading,krizhevsky2009learning}.
Training is done in a sequential way with 20 epochs per task --- in epochs 1-20 we train on MNIST (first task), and on epochs 81-100 we train on CIFAR10 (last task).
All five datasets consist of about 50,000 training images from 10 different classes, but they differ from each other in various ways: black and white vs. RGB, letters and digits vs. vehicles and animals etc.
We use the exact same setup as in \cite{ritter2018online} for the comparison --- LeNet-like \citep{lecun1998gradient} architecture with separated last layer for each task as in CIFAR10/CIFAR100 experiment. Results are reported on Table \ref{table:visiondatasets_results}.
\begin{table*}[t]
\centering
\caption{Accuracy for each task after training sequentially on all tasks. PTL stands for Per-Task Laplace (one penalty per task), AL is Approximate Laplace (Laplace approximation of the full posterior at the mode of the approximate objective) and OL is Online Laplace approximation. Results for SI, PTL, AL and OL are as reported in \cite{ritter2018online}. We highlight the best accuracy in bold.}
\label{table:visiondatasets_results}     
\begin{tabular}{lcccccc}
\multicolumn{1}{l}{}      & \multicolumn{6}{c}{\textbf{Test accuracy [\%] on the end of last task (CIFAR10)}} \\

\multicolumn{1}{l|}{\textbf{Method}} & \multicolumn{1}{c|}{\textbf{Average}} & \textbf{MNIST}                & \textbf{notMNIST}             & \textbf{F-MNIST}         & \textbf{SVHN}             
& \textbf{CIFAR10} \\
\hline
\multicolumn{7}{l}{\rule{0pt}{3ex} \underline{Diagonal methods}}\\
\multicolumn{1}{l|}{FOO-VB}                   &  \multicolumn{1}{c|}{81.37} & 86.42                & 89.23                & 83.05                 & \textbf{82.21}                & \multicolumn{1}{c}{\textbf{65.96}}   \\
\multicolumn{1}{l|}{SGD}                   &  \multicolumn{1}{c|}{69.64} & 84.79                & 82.12                & 65.91                 & 52.31                & \multicolumn{1}{c}{63.08}  \\
\multicolumn{1}{l|}{ADAM}                   &  \multicolumn{1}{c|}{29.67} & 17.39                & 26.26                & 25.02                 & 15.10                & \multicolumn{1}{c}{64.62}        \\
\multicolumn{1}{l|}{SI}                &   \multicolumn{1}{c|}{77.21} & 87.27                & 79.12                & 84.61                & 77.44                & \multicolumn{1}{c}{57.61}                  \\
\multicolumn{1}{l|}{PTL}     &   \multicolumn{1}{c|}{\textbf{82.96}} & \textbf{97.83}                & \textbf{94.73}                & 89.13                & 79.80                & \multicolumn{1}{c}{53.29}                  \\
\multicolumn{1}{l|}{AL}           &   \multicolumn{1}{c|}{82.55} & 96.56                & 92.33                & \textbf{89.27}                & 78.00                & \multicolumn{1}{c}{56.57}                  \\
\multicolumn{1}{l|}{OL}           &   \multicolumn{1}{c|}{82.71} & 96.48                & 93.41                & 88.09                & 81.79                & \multicolumn{1}{c}{53.80}                  \\

\multicolumn{7}{l}{\rule{0pt}{3ex} \underline{Non-Diagonal methods}}\\
\multicolumn{1}{l|}{PTL} &   \multicolumn{1}{c|}{85.32} & 97.85                & \textbf{94.92}       & 89.31                & \textbf{85.75}       & \multicolumn{1}{c}{58.78}                   \\
\multicolumn{1}{l|}{AL} &    \multicolumn{1}{c|}{85.35} & \textbf{97.90}       & 94.88                & 90.08                & 85.24                & \multicolumn{1}{c}{58.63}                   \\
\multicolumn{1}{l|}{OL} &   \multicolumn{1}{c|}{\textbf{85.40}} & 97.17                & 94.78                & \textbf{90.36}       & 85.59                & \multicolumn{1}{c}{\textbf{59.11}}                    \\
    
\end{tabular}
\end{table*}

\clearpage

\bibliography{NECO-20-002-15R1-Ref.bib}
\bibliographystyle{apalike}

\end{document}